\documentclass{article}

\usepackage{amsmath}
\usepackage{graphicx}
\usepackage{subcaption}
\usepackage{hyperref}
\usepackage{cleveref}
\usepackage{algorithm,algorithmic}
\usepackage{tcolorbox}
\usepackage{bbding} 
\usepackage{booktabs}
\usepackage{multirow}
\usepackage[figuresright]{rotating}
\usepackage{longtable}

\usepackage[preprint,nonatbib]{neurips_2025}

\usepackage[utf8]{inputenc}
\usepackage[T1]{fontenc}  
\usepackage{hyperref}       
\usepackage{url}           
\usepackage{booktabs}       
\usepackage{amsfonts}       
\usepackage{nicefrac}     
\usepackage{microtype}     
\usepackage{xcolor}        
\usepackage{enumitem}
\usepackage{booktabs}
\usepackage{subcaption}
\usepackage{tabularx, booktabs}

\DeclareSymbolFont{extraup}{U}{zavm}{m}{n}
\DeclareMathSymbol{\varheart}{\mathalpha}{extraup}{86}
\DeclareMathSymbol{\vardiamond}{\mathalpha}{extraup}{87}
\DeclareMathSymbol{\varclubsuit}{\mathalpha}{extraup}{88}

\tcbset{
  mybox/.style={
    colback=white, colframe=black,
    fonttitle=\bfseries,
    left=1mm, right=1mm, top=1mm, bottom=1mm,
    boxsep=1mm, arc=1mm,
  }
}

\title{Multimodal Conditioned Diffusive \\Time Series Forecasting}

\author{%
  Chen Su$^{{\spadesuit}}$, \hspace{0.1cm}
    Yuanhe Tian$^{\varheart}$, \hspace{0.1cm}
    Yan Song$^{{\spadesuit}*}$
    \\
    $^{\spadesuit}$University of Science and Technology of China \\
    $^{\varheart}$University of Washington
    \\
    $^{\spadesuit}$\texttt{suchen4565@mail.ustc.edu.cn} \hspace{0.1cm}
    $^{\varheart}$\texttt{yhtian@uw.edu} \hspace{0.1cm}
    $^{\spadesuit}$\texttt{clksong@gmail.com}
}

\begin{document}

\renewcommand{\thefootnote}{\fnsymbol{footnote}}
\footnotetext[1]{Corresponding author.}

\renewcommand{\thefootnote}{\arabic{footnote}}

\maketitle

\begin{abstract}
Diffusion models achieve remarkable success in processing images and text, and have been extended to special domains such as time series forecasting (TSF).
Existing diffusion-based approaches for TSF primarily focus on modeling single-modality numerical sequences, overlooking the rich multimodal information in time series data.
To effectively leverage such information for prediction, we propose a multimodal conditioned diffusion model for TSF, namely, MCD-TSF,
to jointly utilize timestamps and texts as extra guidance for time series modeling, especially for forecasting.
Specifically, 
Timestamps are combined with time series to establish temporal and semantic correlations among different data points when aggregating information along the temporal dimension.
Texts serve as supplementary descriptions of time series' history, and adaptively aligned with data points as well as dynamically controlled in a classifier-free manner.
Extensive experiments on real-world benchmark datasets across eight domains demonstrate that the proposed MCD-TSF model achieves state-of-the-art performance.
\footnote{The code is available at \url{https://github.com/synlp/MCD-TSF}.}

\end{abstract}

\section{Introduction}

Time series forecasting (TSF) is a critical task that leverages historical data to predict future time series, enabling proactive decision-making across a wide range of industries. By extracting meaningful insights from temporal dependencies, TSF not only enhances operational efficiency but also mitigates risks, optimizes resource allocation, and supports long-term strategic planning. There are many real-world scenarios TSF has been applied to, such as financial analysis \cite{financialAPP1, financialAPP2, financialAPP3}, traffic flow estimation \cite{trafficAPP1, trafficAPP2, trafficAPP3}, climate modeling \cite{climateAPP1, climateAPP2, climateAPP3}, and energy consumption prediction \cite{energyApp1, energyApp2, energyApp3}, etc.

With the rise of deep learning, the TSF research shifts from traditional statistical models \cite{intro:arima1, intro:arima2} toward neural networks \cite{intro:neural, intro:neural1} that exhibit large model capacity, strong scalability, and an end-to-end training paradigm.
These models take a time series, i.e., a sequence of values arranged in chronological order, as input and predict values at future times.
Specifically, convolutional models \cite{cnn-model1, cnn-model2} leverage kernel and window operations to extract local patterns from time series, whereas they are difficult to obtain the relations among data points at different times.
Recurrent models \cite{rnn-model1, rnn-model2} capture temporal dependencies via self-repeating processes, with their deficits in capturing long-distance features.
Transformer-based models \cite{transformer-model1, transformer-model2, transformer-model3} utilize self-attention mechanisms to model cross dependencies among different data points and generally achieve better performance than the convolutional and recurrent models. 
In recent years, pretrained models, especially large language models (LLMs), have demonstrated powerful data processing capabilities \cite{gpt4,llama3,chimed,qwen2}, and TSF datasets are developed with extra annotations (such as timestamps and textual descriptions) available along with the time series \cite{time-mmd}.
Therefore, researchers begin to use LLMs to borrow their knowledge from pretrained data and handle rich inputs of time series such as texts \cite{llm-model1, llm-model2, timellm}.
Straightforwardly, to leverage the LLMs' inherent capabilities for text-processing, studies \cite{llmtime, test, chattime} generally convert time series directly into text for the model to process, and thus, however, 
lose the continuous numerical characteristics inherent in time series.
Meanwhile, these approaches treat the time series as fixed input, so that slight variations in the input cause significant impact on the prediction results \cite{llm_sense1, llm_sense2}.
Consider that a key characteristic of time series data is its presence of a degree of randomness and volatility, such variations should have limited impact on future outcomes.
So that
the gap between the aforementioned approaches and the characteristics of time series 
leads to a significant bottleneck and calls for solutions
that are able to address this randomness, such as probabilistic models.
In doing so, recent studies \cite{TimeGrad, CSDI, d3vae, MG-TSD, mr-Diff} utilize diffusion networks
for TSF. 
They establish probabilistic relational mappings between historical time series and future data points through parameterized Markov processes.
Most of them
only use historical time series as input, extracting features from the series itself to condition and guide the diffusion process.
Although others \cite{VE_diff_TSF} generate text descriptions of statistical features (such as the median and trend) of the time series and use these descriptions to enhance the understanding of the time series, they remain restricted by historical time series and textual data, with less attention paid to other types of information, such as timestamps,
which encode structural and hierarchical temporal relationships (e.g., day, week, and month) among successive data points, have been demonstrated to provide essential hints for modeling time-related data \cite{glaff, timelinear}.
Therefore, with the aforementioned limitations from existing studies, it is expected to develop a probabilistic approach to integrate multimodal information, particularly able to process timestamps, thereby enhancing the understanding and forecasting of time series data.

In this paper,
we propose a multimodal conditioned diffusion model for TSF, namely,
MCD-TSF,
which leverages multiple input sources from different modalities such as timestamps and texts to enhance the standard time series input, so as to help the diffusion model to better perform TSF.
Specifically, at each diffusion step, we firstly encode the historical time series, timestamps and texts,
and then fusing their embeddings,
where we integrate timestamp embeddings into time-series embeddings to enhance the understanding for each data point, and then further integrate such enhanced
representations with the textual embeddings,
so as to progressively guide the denoising process.
Moreover, to mitigate the influence of the potentially irrelevant textual content, we employ {adaptive balancing technique named classifier-free guidance (CFG) \cite{cfg}} 
to dynamically fuse the time series representations by two settings: timestamp only or using both timestamp and text, providing better robustness of our approach under variant inputs, e.g., missing the text part.
As a result,
MCD-TSF
allow the TSF task not only leverages the probabilistic nature of diffusion models to accurately capture the inherent randomness of time-series data, but also achieves a deeper and more nuanced understanding of temporal patterns through dynamic fusion of multimodal information. 
Extensive experiments on multimodal datasets covering eight real-world domains, including transportation, agriculture, weather, economics, and energy,
demonstrate the superior generalized performance of our approach, which outperforms strong baselines and existing studies.

\section{Related Work}

\subsection{Diffusion Models for Time Series Modeling}

Diffusion models \cite{DDPM,diffusion,ddim} are probabilistic neural approaches widely used for {many tasks (e.g., image generation)}, and have recently been applied to time series modeling \cite{diff_tsf1,diff_tsf2,TimeGrad,d3vae,RATD}.
Many diffusion-based studies \cite{CSDI,diff4interpo1} focus on time series interpolation, where they learn a stochastic mapping from the given time series to the missing values based on their left and right time contexts.
{These studies} conventionally treat historical numerical sequences as denoising conditions to guide the model in reconstructing the time series from Gaussian noise. 
{
However, standard diffusion models overlook inherent temporal patterns and multi-scale characteristics in the data, limiting prediction accuracy.
}
To enhance the accuracy of time series prediction, recent studies leverage {signal processing
} techniques such as season-trend decomposition \cite{mr-Diff}, multi-resolution analysis \cite{MG-TSD, mr-Diff}, and spectral analysis \cite{diffusion-ts} to extract semantic information from time series data.
There are also studies \cite{CSDI, TimeDART} utilizing self-supervised pre-training to learn high-quality representations of time series {to improve the understanding of historical time series and thus obtain better TSF performance}.
Although these approaches provide promising solutions to improving time series modeling, they pay insufficient attention to leveraging extra data or inputs from other modalities that provides important information for time series understanding.
Meanwhile, existing studies {focus on time series interpolation or other time series tasks and}, to some extent, merely touch the forecasting task.
Compared to existing studies, we apply diffusion models focusing on TSF that predicts the time series based solely on the left context and utilize multimodal inputs such as timestamps and text to faciliate the task.

\begin{figure*}[t]
\begin{center}
\centerline{\includegraphics[width=\textwidth, trim=0 10 0 0]{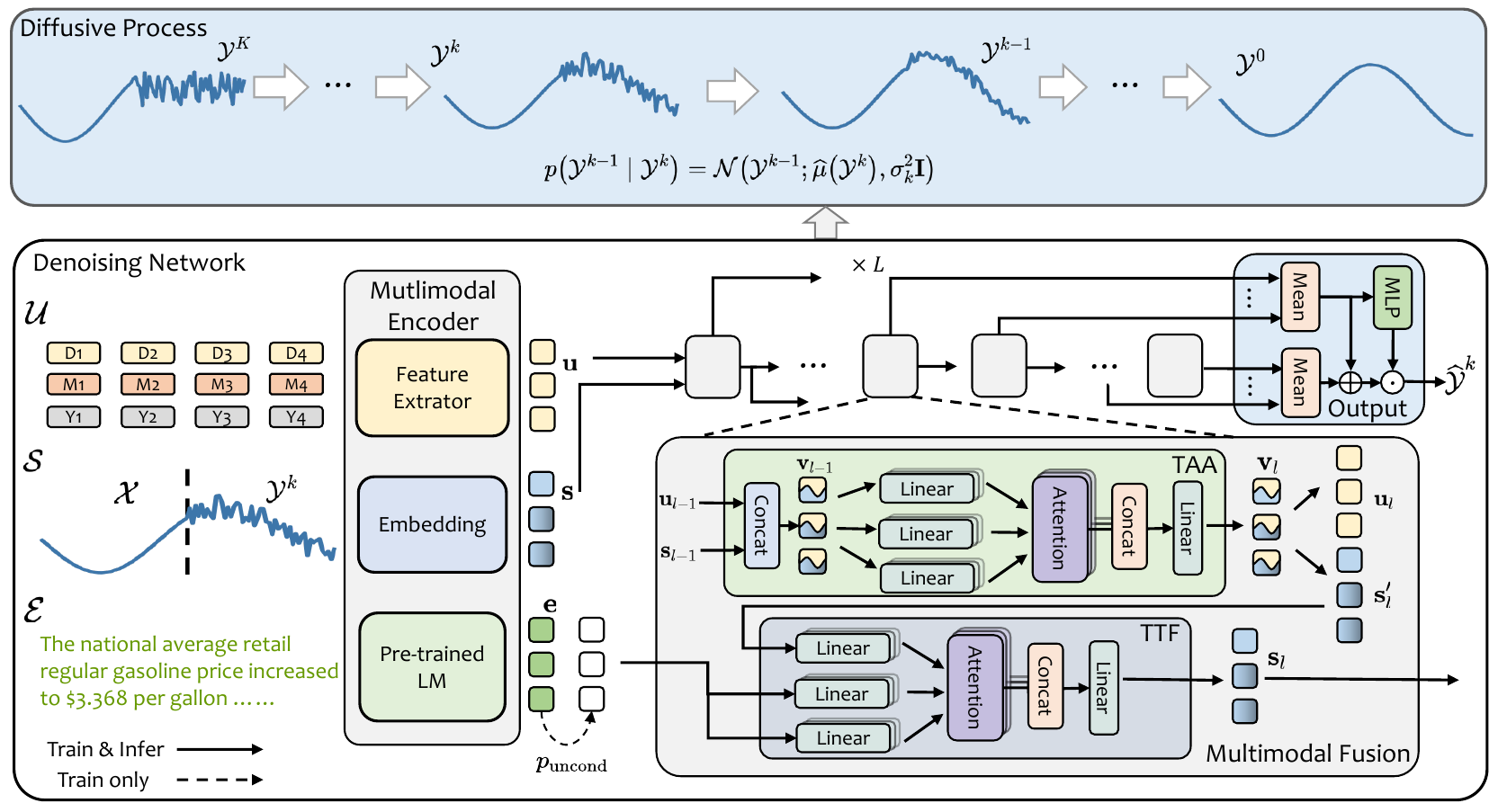}}
\caption{{The overall architecture of the proposed MCD-TSF. 
Our approach consists of four components: the diffusion framework; a multimodal encoder, multimodal fusion module, and an output layer, which are presented at the top, bottom-left, bottom-center, and bottom-right parts of the figure, respectively. 
The multimodal fusion component internally contains the TAA and TTF modules. 
}
}
\label{Overall Architecture}
\end{center}
\vskip -0.3in
\end{figure*}

\subsection{Multimodal Time Series Forecasting}

Another research line of this work is multimodal TSF, where
the models incorporate heterogeneous inputs beyond time series data.
Most multimodal studies \cite{patchTST, dlinear, informer, autoformer, fedformer} utilize text as the additional modality to enhance time series inputs.
To fuse text information into the time series for TSF, current studies \cite{promptcast, unitime, autotimes, timerag} employ LLMs, which are used as backbones and concatenate text embeddings with time series embeddings to form the new input, with
the final predicted value sequence generated by mapping the model's hidden states through linear layers. 
In doing so, they
rely on converting time series into specialized textual data comprehensible to LLMs, thereby losing the numerical information in the original time series data.
Some studies \cite{s2ip-llm, ltsm-bundle} directly concatenate time series embeddings with text embeddings as model input and encode them to predict the future values.
To further improve model performance,
{Liu et al. (2024)}
\cite{time-mmd} combine LLM-encoded textual features with single-modal TSF model outputs through learnable linear weighting to produce future values.
{Zhou et al. (2025)}
\cite{textfusionHTS} employ cross-attention to enhance cross-modal fusion, subsequently completing predictions via feed-forward networks.
Although these approaches achieve fine-grained text-time series fusion, they fail to {control the influence of text on time series modeling}, 
leading to textual noise-induced degradation of forecasting accuracy.
{
However, relying on full‐scale LLMs for text encoding often incurs excessive compute and memory costs, and is not strictly necessary for TSF tasks where the textual inputs are relatively short and domain‐specific.
Therefore, there are studies \cite{time-mmd} that utilize relatively small pre-trained language models (e.g., BERT \cite{bert}) to encode text and fuse the resulting text representations to time series presentations, and obtain promising TSF performance.
}
There are some recent studies \cite{glaff,timelinear} that utilize timestamps to enhance TSF, where the timestamps are independently predicted in their models and then combined with the historical time-series-based prediction through weighting to predict the final future values. 
They focus on integrating global patterns from timestamps while neglecting their roles in representing temporal hierarchical relationships among data points.
{
These approaches are inherently deterministic and therefore cannot quantify the uncertainty in their predictions, nor do they provide a probabilistic view of future trajectories, which is a significant limitation when time series data contain volatility and noise.
}
Compared with existing approaches, our MCD-TSF is designed on diffusion models that {extracts structural temporal cues from timestamps and semantic context from text, then jointly fuses them with the numerical time series to perform uncertainty-aware forecastings.}

\section{The Approach}
The overall architecture of MCD-TSF is illustrated in Figure \ref{Overall Architecture}.
Given a time series {$\mathcal{X}=\{\mathbf{x}_1 \cdots \mathbf{x}_H\}$ of $H$ values (the $h$-th value is $\mathbf{x}_h$)} with its corresponding timestamps $\mathcal{T}$ and related texts $\mathcal{E}$, as well as the future timestamps $\mathcal{T}'$, MCD-TSF progressively predicts the future {time series $\mathcal{Y}=\{\mathbf{x}_{H+1} \cdots \mathbf{x}_{H+F}\}$ with $F$ values} {at future times $\mathcal{T}'$}
from a Gaussian noise vector $\mathbf{n}$ through a $K$-step denoising process $\{\mathcal{Y}^K, \mathcal{Y}^{K-1}, ..., \mathcal{Y}^k, \mathcal{Y}^{k-1}, ..., \mathcal{Y}^0 \}$, where $\mathcal{Y}^k$ is the denoised output {(state)} of the diffusion model at the $k$-th step with the initial $\mathcal{Y}^K$ equaling to $\mathbf{n}$ and the final $\mathcal{Y}^0$ referring to $\mathcal{Y}$.
The denoising process at each step $k$ is performed by a denoising network $f_\theta$,
which takes $\mathcal{Y}^k$, $\mathcal{T}'$, and multimodal inputs (including $\mathcal{X}$, $\mathcal{T}$, and $\mathcal{E}$) to predict $\mathcal{Y}^{k-1}$, where the timestamp and text are used {one after the other} to enhance the time series representation for better performing the denoising process.
By iterating this denoising process $K$ times, we obtain an estimate of the future value {$\widehat{\mathcal{Y}}$ (which is identical to $\mathcal{Y}^0$ in the denosing process)} in the time series.
In the following subsections, we firstly present the principle of diffusive modeling for TSF, then describe the multimodal conditioned diffusive process, and finally illustrate the details of our MCD-TSF model.

\subsection{Diffusive Modeling for TSF}

Conventional diffusion models 
consist of a forward noise-addition and a reverse denoising process.
In the forward process, the diffusion model iteratively adds Gaussian noise to the target \(\mathcal{Y}\) in the sequence of states in the denoising process \(\{\mathcal{Y}^k\}_{k=1}^K\).
At the \(k\)-th diffusion step, the state \(\mathcal{Y}^k\) is drawn from the target \(\mathcal{Y}\) as follows:
\begin{equation} \label{eq:forward}
    p\bigl(\mathcal{Y}^k \mid \mathcal{Y}\bigr)
    = \mathcal{N}\!\Bigl(\mathcal{Y}^k \;;\;\mathbf{\mu} = \sqrt{\bar{\alpha}_k}\,\mathcal{Y},\;\mathbf{\Sigma} = (1 - \bar{\alpha}_k)\, \mathbf{I}\Bigr)
\end{equation}
where \(p(\cdot\mid\cdot)\) denotes a conditional probability density function; \(\mathcal{N}(x;\,\mu,\Sigma)\) refers to the multivariate Gaussian distribution with mean vector \(\mathbf{\mu}\) and covariance matrix \(\mathbf{\Sigma}\); and \(\mathbf{I}\) is the identity matrix whose dimensionality matches that of \(\mathcal{Y}\).
Herein, \(\bar{\alpha}_k = \prod_{i=1}^k \alpha_i\) is the cumulative signal retention factor up to step \(k\), with each \(\alpha_i \in (0,1)\) controlling the proportion of signal preserved (and thus the noise level \(1 - \alpha_i\)) at step \(i\).

In the reverse process, the diffusion model aims to recover the target \(\mathcal{Y}\) from pure noise \(\mathcal{Y}^K\) through a parameterized denoising distribution:
\begin{equation} \label{eq:reverse}
    p(\mathcal{Y}^{k-1} \mid \mathcal{Y}^k)
    = \mathcal{N}\bigl(\mathcal{Y}^{k-1}; \widehat{\mu}(\mathcal{Y}^k),\,\sigma_k^2 \mathbf{I}\bigr)
\end{equation}
where \(\sigma_k^2\) is the variance associated with diffusion step \(k\) and \(\widehat{\mu}(\mathcal{Y}^k)\) is 
computed by
\begin{equation}
\label{sample2}
\widehat{\mu}(\mathcal{Y}^k)=
    \frac{\sqrt{\bar\alpha_{k}}\left(1-\bar{\alpha}_{k-1}\right)}{\sqrt{\bar\alpha_{k-1}}(1-\bar{\alpha}_{k})} \mathcal{Y}^k
    + \frac{\bar\alpha_{k-1}-\bar\alpha_{k}}{\sqrt{\bar\alpha_{k-1}}(1-\bar\alpha_k)}  \widehat{\mathcal{Y}}^k
\end{equation}
where $\widehat{\mathcal{Y}}^k$ is obtained through a diffusion network $f_\theta$ parameterized by \(\theta\) via 
\begin{equation} \label{eq:y-hat}
\widehat{\mathcal{Y}}^k=f_{\theta}(\mathcal{Y}^{k})
\end{equation}
By randomly sampling diffusion steps \(k\) from a discrete uniform distribution and minimizing the following loss, the parameters \(\theta\) of the denoising network are learned by
\begin{equation} \label{eq:loss}
    \mathcal{L}(\theta)
    = \mathbb{E}_{k \sim \mathrm{Uniform}\{1,\dots,K\}}
      \Bigl[\bigl\|\,f_{\theta}(\mathcal{Y}^k)\,-\,\mathcal{Y}^*\bigr\|_{2}^{2}\Bigr]
\end{equation}
where \(\mathcal{L}(\theta)\) is the scalar expected mean squared error loss; \(\mathbb{E}[\cdot]\) denotes expectation over the random variable \(k\); \(\mathrm{Uniform}\{1,\dots,K\}\) denotes the discrete uniform distribution on the set \(\{1,\dots,K\}\); $\mathcal{Y}^*$ is the gold standard;
and \(\|\mathbf{v}\|_2 = \sqrt{\sum_{i=1}^d v_i^2}\) is the Euclidean norm of vector \(\mathbf{v}\in\mathbb{R}^d\).

When applying the diffusion model on TSF, it utilizes the same forward process as Eq. (\ref{eq:forward}) and conditions the reverse process on historical inputs to help the denoising process.
The reverse denoising distribution in Eq. (\ref{eq:reverse}) is then formulated as
\begin{equation}
    p(\mathcal{Y}^{k-1} \mid \mathcal{Y}^k, \mathcal{C})
    = \mathcal{N}\bigl(\mathcal{Y}^{k-1}; \widehat{\mu}(\mathcal{Y}^k, \mathcal{C}),\,\sigma_k^2 \mathbf{I}\bigr)
\end{equation}
where the network \(f_\theta\) takes \(\mathcal{Y}^k\) and condition \(\mathcal{C}\) (i.e., the historical inputs), outputs the estimate \(\widehat{\mu}\) of the next denoised state.
Meanwhile, the model still follows Eq. (\ref{sample2}) to compute $\widehat{\mu}(\mathcal{Y}^k, \mathcal{C})$ with Eq. (\ref{eq:y-hat}) formulated as
\begin{equation}
    \widehat{\mathcal{Y}}^k=f_{\theta}(\mathcal{Y}^{k}, \mathcal{C})
\end{equation}
During training, 
one minimizes the following conditional loss:
\begin{equation}
    \mathcal{L}(\theta)
    = \mathbb{E}_{k \sim \mathrm{Uniform}\{1,\dots,K\}}
      \Bigl[\bigl\|\,f_{\theta}(\mathcal{Y}^k,\mathcal{C})\,-\,\mathcal{Y}^*\bigr\|_{2}^{2}\Bigr]
\end{equation}
so that the diffusion model learns to produce probabilistic forecasts of future values for TSF on the condition $\mathcal{C}$.

\subsection{Multimodal Conditioned Diffusive Process for TSF}

With multimodal input, the condition variable $C$ consists of not only the historical time series $\mathcal{X}$, but also other sources such as timestamps $\mathcal{U}=[\mathcal{T}, \mathcal{T}']$ (which includes the historical timestamps $\mathcal{T}$ that correspond to $\mathcal{X}$ and the future timestamp $\mathcal{T}'$), and textual descriptions $\mathcal{E}$, etc.
The forward diffusion process follows the standard procedure in Eq. (\ref{eq:forward}).
In the reverse diffusion process, the denoising network uses a conditioned Gaussian distribution to iteratively recover the future time series by
\begin{equation}\label{sample1}
p\bigl(\mathcal{Y}^{k-1}\mid \mathcal{Y}^k, \mathcal{X}, \mathcal{U}, \mathcal{E} \bigr)
= \mathcal{N}\!\bigl(\mathcal{Y}^{k-1};\,\widehat{\mu}_\theta(\mathcal{Y}^k,\mathcal{X}, \mathcal{U}, \mathcal{E}),\,\sigma_k^2 I\bigr)
\end{equation} 
where $\widehat{\mu}_\theta(\mathcal{Y}^k,\mathcal{X}, \mathcal{U}, \mathcal{E})$ is computed by Eq. (\ref{sample2}) and Eq. (\ref{eq:y-hat}) is formulated by
\begin{equation} \label{eq:ours-y-hat}
    \widehat{\mathcal{Y}}^k=f_{\theta}(\mathcal{Y}^{k}, \mathcal{X}, \mathcal{U}, \mathcal{E})
\end{equation}
$f_{\theta}$
is then trained by minimizing the following conditional mean squared error loss
\begin{equation}
\mathcal{L}(\theta)
= \mathbb{E}_{k\sim \mathrm{Uniform}\{1,\dots,K\}}
\Bigl[\|\,f_\theta(\mathcal{Y}^k, \mathcal{X}, \mathcal{U}, \mathcal{E})\,-\,\mathcal{Y}\|_2^2\Bigr]
\end{equation}
so that
the model learns to integrate both timestamp and text information to enhance the timeseries representation for a better denoising process.

As for the text part, since textual information often serves as a coarse-grained supplement for time series data, with its effectiveness easily influenced by text quality across domains, it is necessary to design a special mechanism to control the model's reliance on textual information during the inference process to balance the enhancement from timestamps or texts.
In doing so, CFG,
which is
experimentally validated in conditional image generation domains and demonstrates its capability to balance specificity and diversity in image generation \cite{cfg},
is utilized to perform our mechanism.
In detail, we adjust the prediction \(\widehat{\mathcal{Y}}\) of the denoising network through a linear combination of text-conditioned and non-text-conditioned (i.e., input text is set to an empty string denoted by $\emptyset$) components, which is represented by
\begin{equation}
\label{cfg:inference}
{\widehat{\mathcal{Y}}^{k}_w}=f_\theta(\mathcal{Y}^k,\mathcal{X},\mathcal{U},\emptyset)+w\left(f_\theta(\mathcal{Y}^k,\mathcal{X},\mathcal{U},\mathcal{E})-f_\theta(\mathcal{Y}^k,\mathcal{X},\mathcal{U},\emptyset)\right)
\end{equation}
where \(w\) is a hyperparameter that controls the model's reliance on textual information
and {$\widehat{\mathcal{Y}}^k_w$ replace ${\widehat{\mathcal{Y}}}^k$ in Eq. (\ref{eq:ours-y-hat})} during the diffusion process.
In practice,
when training the model with and without text conditions (i.e., $f_\theta(\mathcal{Y}^k,\mathcal{X},\mathcal{U},\mathcal{E})$ and $f_{\theta}(\mathcal{Y}^k,\mathcal{X},\mathcal{U},\emptyset)$, respectively), we follow \cite{cfg} to randomly mask the text conditions with a certain probability $p_{uncond}$,
so as to enable simultaneous training of both conditional and unconditional cases for better generalization.

\subsection{The MCD-TSF Model}

The MCD-TSF consists of three core modules: a multimodal encoder, a stack of Transformer-based fusion layers, and an output layer for adaptive feature fusion.

The multimodal encoder transforms the {combined time series $\mathcal{S}=[\mathcal{X},\mathcal{Y}^k]$}, timestamps $\mathcal{U}$, and text $\mathcal{E}$ into feature embeddings $\mathbf{s}$, $\mathbf{u}$, and $\mathbf{e}$, respectively.  
Specifically, $\mathbf{s}$ is obtained by applying a $1\times1$ convolutional layer to extract numerical features from $\mathcal{S}$.  
There is a
timestamp feature extractor, which firstly disentangles structural attributes (e.g., \texttt{DayOfWeek}, \texttt{DayOfMonth}, \texttt{DayOfYear}) from the raw timestamps, normalizes each to $[-0.5,0.5]$, which results in vectors such as $[0.166,0.3,-0.02]$, and then encodes this vector via a $1\times1$ convolution to produce the embedding $\mathbf{u}$.
The text input $\mathcal{E}$ is tokenized and encoded using a frozen pretrained {language model (e.g., BERT \cite{bert})}, with its final hidden states serving as the textual embedding $\mathbf{e}$.

The Transformer-based fusion module contains $L$ layers, where each layer consists of Timestamp-assisted attention (TAA) and text-time series fusion (TTF) to progressively enhance the time-series features.\footnote{{The effectiveness of firstly performing TAA and then utilizing TTF is supported by the experimental results in Appendix \ref{app:exp model order}.}}
At layer $l$, each fusion module takes as input the previous layer’s time-series representation $\mathbf{s}_{l-1}$, timestamp representation $\mathbf{u}_{l-1}$, and the static text representation $\mathbf{e}$, and outputs the timestamp-enhanced sequence representation $\mathbf{s}_l$, the updated timestamp representation $\mathbf{u}_l$.
Specifically, in TAA, we stack $\mathbf{s}_{l-1}$ and $\mathbf{u}_{l-1}$ to obtain a joint representation $\mathbf{v}_{l-1} = [\mathbf{s}_{l-1}; \lambda \cdot \mathbf{u}_{l-1}]$ (where $\lambda$ is the weight to control the contribution of the timestamp information).
We process $\mathbf{v}_{l-1}$ using a multi-head self-attention and obtain $\mathbf{v}_l$, which is also formulated as $\mathbf{v}_l=[\mathbf{s}'_{l}; \mathbf{u}_{l}]$ with $\mathbf{s}'_{l}$ and $\mathbf{u}_{l}$ corresponding to $\mathbf{s}_{l-1}$ and $\mathbf{u}_{l-1}$, respectively.
In TTF, we use text to enhance features in the time series through cross-attention, where the timestamp-enhanced time series features $\mathbf{s}'_l$ serve as queries while the textual features $\mathbf{e}$ act as keys and values.
Therefore,
we obtain the augmented time series features $\mathbf{s}_l$ that carry the numerical semantics of the time series, temporal structural information from timestamps, and complementary information from textual descriptions.
Finally, the computed $\mathbf{u}_{l}$ and $\mathbf{s}_l$ are fed into the next layer for further processing.

The output layer performs two predictions separately from the layer-wise text-timestamp features \(\{\mathbf{s}_l\}_{l=1}^L\) and timestamp features \(\{\mathbf{u}_l\}_{l=1}^L\), and then adaptively fuse them to obtain the final output.
Specifically, we firstly apply mean pooling to $\{\mathbf{s}_l\}_{l=1}^L$ and $\{\mathbf{u}_l\}_{l=1}^L$ to compute their averaged representations $\mathbf{s}_{\mathrm{avg}}$ and $\mathbf{u}_{\mathrm{avg}}$ by
\begin{equation}
  \mathbf{s}_{\mathrm{avg}}
  = \frac{1}{\sqrt{L}}\sum_{l=1}^L \mathbf{s}_{l},\quad
  \mathbf{u}_{\mathrm{avg}}
  = \frac{1}{\sqrt{L}}\sum_{l=1}^L \mathbf{u}_{l}
\end{equation}
Then, two multi-layer $1\times1$ convolutional heads are applied to produce the time series-based and timestamp-based predictions
(which are denoted as {\((\widehat{\mathcal{X}}_s,\widehat{\mathcal{Y}}_s)\) and \((\widehat{\mathcal{X}}_u,\widehat{\mathcal{Y}}_u)\), respectively) }by:
{
\begin{equation}
  (\widehat{\mathcal{X}}_s,\widehat{\mathcal{Y}}_s)
  = \mathrm{Conv}_s(\mathbf{s}_{\mathrm{avg}}),\quad
  (\widehat{\mathcal{X}}_u,\widehat{\mathcal{Y}}_u)
  = \mathrm{Conv}_u(\mathbf{u}_{\mathrm{avg}})
\end{equation}}
where {$\widehat{\mathcal{X}}_s$ and $\widehat{\mathcal{Y}}_s$} are the time series-based predictions of the historical time series $\mathcal{X}$ and {$\mathcal{Y}$}, respectively;
{$\widehat{\mathcal{X}}_u$ and $\widehat{\mathcal{Y}}_u$ are the timestamp-based predictions according to the historical timestamps $\mathcal{T}$ and the future timestamp $\mathcal{T}'$.}
Finally, a learnable multi-layer perceptron (MLP) computes fusion weights \(\mathbf{W}\) from the differences between {$\widehat{\mathcal{X}}_u$ and $\mathcal{X}$} by
\begin{equation}
  \mathbf{W} = \mathrm{MLP}(\widehat{\mathcal{X}}_u - \mathcal{X})
\end{equation}
which is used to {fuse $\widehat{\mathcal{Y}}_s$ and $\widehat{\mathcal{Y}}_u$} by
{
\begin{equation}
  \widehat{\mathcal{Y}}_{k-1}
  = \sum \mathbf{W} \odot \bigl(\widehat{\mathcal{Y}}_s \oplus \widehat{\mathcal{Y}}_u\bigr)
\end{equation}
}
where \(\odot\) and \(\oplus\) stand for Hadamard product and concatenation operations, respectively, and the resulting {$\widehat{\mathcal{Y}}^{k-1}$} is fed into the next diffusion step for denoising to {$\mathcal{Y}^0$}.

\begin{table}[t]
    \centering
    \caption{Statistics of the dataset. Frequency denotes the temporal recording frequency of the time series; number of samples indicates the quantity of time series samples; text coverage represents the ratio of the number of texts to the number of temporal samples in the domain.
    ``Arg'', ``Cli'', ``Eco'', ``Ene'', ``Env'', ``Hea'', ``Soc'', and ``Tra'' are abbreviations for agriculture, climate, economy, energy, environment, health, social good, and traffic, respectively.
    \label{tab:dataset}
    }
\resizebox{\linewidth}{!}{
    \begin{tabular}{l|cccccccc}
    \toprule[2pt]
        ~ &  \textbf{Agr} & \textbf{Cli} & \textbf{Eco} & \textbf{Ene} & \textbf{Env} & \textbf{Hea} & \textbf{Soc} & \textbf{Tra} \\ 
    \midrule
        Frequency & Monthly & Weekly & Monthly & Weekly & Daily & Weekly & Monthly & Monthly \\ 
        Number of Samples & 496 & 496 & 423 & 1479 & 11102 & 1389 & 900 & 531 \\ 
        Text Coverage & 13.3\% & 100\% & 85.4\% & 37.4\% & 4.2\% & 54.9\% & 41.3\% & 43.4\% \\ 
    \bottomrule[2pt]
    \end{tabular}}
\end{table}

\section{Experiment Setting}
\subsection{Dataset}
We follow existing studies \cite{time-mmd} to utilize the text-time series benchmark datasets that cover eight domains, namely, \textit{agriculture}, \textit{climate}, \textit{economy}, \textit{energy}, \textit{environment}, \textit{health}, \textit{social good}, and \textit{traffic}, respectively.
We normalize the data to regularize it into a distribution with a mean of 0 and a standard deviation of 1.
Following existing studies \cite{autoformer, fedformer}, the dataset is then partitioned chronologically into training, validation, and test sets in a 7:1:2 splitting.
Table \ref{tab:dataset} summarizes the key statistics of all domains, including frequency, number of samples, and text coverage.

\subsection{Baselines and Comparing Approaches}
\label{sec:baseline}

To evaluate our proposed MCD-TSF, we compare five baseline models illustrated as follows.
\begin{itemize}[leftmargin=1.5em, labelsep=0.8em]
    \item DIFF: the standard diffusion model with time series as the only input.
    \item DIFF+TAA: it extends the DIFF model by incorporating timestamp input into the time series with the TAA module.
    \item DIFF+TTF: it extends the DIFF model by adding textual input to the time series through the TTF module.
    \item DIFF+TAA-T: it builds upon DIFF+TAA with additional textual input and the MM-TSF \cite{time-mmd} text integration module. MM-TSF maps the hidden states of the last layer of the LLM into predictions via a linear layer. These predictions are then adaptively weighted and integrated with the forecasting predictions derived from historical time series data.
    \item DIFF+TTF-T: it enhances DIFF+TTF by including timestamp input and the TimeLinear \cite{timelinear} timestamp integration module. Herein, TimeLinear processes the input timestamps through multiple linear layers and $1\times1$ convolutional layers, mapping them into predictions, which are then combined with weighted predictions obtained from historical time series.\footnote{Following the convention, our MCD-TSF is equivalent to DIFF+TAA+TTF.}
\end{itemize}

To further demonstrate the superiority of MCD-TSF, we also conduct comparisons with a wide range of existing models that are summarized as follows.
\begin{itemize}[leftmargin=1.5em, labelsep=0.8em]
    \item Transformer-based models: Reformer \cite{reformer}, Autoformer \cite{autoformer}, FEDformer \cite{fedformer}, PatchTST \cite{patchTST}, and HCAN (integrated with Informer) \cite{hcan};
    \item MLP-based models: FILM \cite{film}, DLinear \cite{dlinear}, and TimeMixer++ \cite{timemixer++};
    \item State space models: Timemachine \cite{timemachine}; 
    \item Probabilistic models: CSDI \cite{CSDI} and D3VAE \cite{d3vae};
    \item LLM-based models: FPT \cite{fpt};
    \item Text-integrated models: TimeLLM \cite{timellm} and MM-TSF (integrated with Informer) \cite{time-mmd}; 
    \item Timestamp-integrated models: GLAFF (integrated with DLinear) \cite{glaff} and TimeLinear \cite{timelinear}.
\end{itemize} 
Note that, while CSDI is a self-attention-based diffusion model for TSF that shares similarities with our approach, it solely incorporates temporal sequence inputs and lacks the capability to process multimodal data.
GLAFF similarly employs attention to handle temporal-stamp features, but adopts a separate processing paradigm where timestamps and time series are modeled independently when capturing temporal relationships.
For all comparing models, we use their official implementations.

\subsection{Implementation Details}
\label{sec:implementation}

We pre-process the dataset based on the following configurations.
According to \cite{time-mmd}, the settings of historical time series length and prediction horizon are determined based on the time series frequency.
For the time series part, to accommodate them with varying recording frequencies (i.e., monthly, weekly, and daily), we define a historical time series length as 36 months for monthly frequency, 96 weeks for weekly frequency, and 336 days for daily frequency.
Meanwhile, the prediction horizon is also frequency-dependent, with three distinct lengths configured for each frequency to more comprehensively evaluate the model's short-, medium-, and long-term forecasting capabilities.
%
For example, we predict the future values at 6, 12, and 18 months for monthly frequency, at 12, 24, and 48 weeks for weekly frequency series, and at 48, 96, and 192 days for daily frequency series.
The configuration of these length settings follows existing TSF studies \cite{autoformer, fedformer} while incorporating refinements based on the following insights: The minimum forecast length should exceed the temporal granularity of the time series recording interval by at least one hierarchical level to better evaluate model capabilities in capturing long-term dependencies.
For instance, with monthly recorded time series (where the recording interval is measured in months), the minimum forecast length should be no less than 6 months. 
Concurrently, the historical sequence length should be at least twice the forecast length to ensure sufficient historical context supports the predictions.
For the timestamp input, their feature dimension aligns with their frequency, e.g., for monthly, weekly, and daily frequencies, the dimensions are set to three, two, and one, respectively.\footnote{{The detailed timestamp encoding process is presented in Appendix \ref{app:encoding of timestamp}.}}
For the text input,
we concatenate all text corresponding to the 36 intervals preceding the end date of the historical time series. 
For instance, in a daily frequency time series where each interval represents a day, we aggregate texts spanning the 36-day period prior to the series' end date into a comprehensive final text, which subsequently serves as the associated text for that particular historical time series.\footnote{{The details of how we obtain the text are illustrated in Appendix \ref{app:text input}.}}

For model architecture, {considering pre-trained language models \cite{bert,zen} have achieved outstanding performance for text modeling with significantly lower computational cost than LLMs}, we employ the pre-trained {BERT-base} \cite{bert} as the text encoder following its default settings, i.e., 12 layers of Transformer with 768-dimensional hidden states.
We use 6 multimodal fusion modules, where both TAA and TTF are implemented using the standard multi-head self-attention.
The hidden layer dimension of the model is set to 64.
For the timestamp weight $\lambda$, the default value is set to 1.
For the textual guidance strength \( w \), the default value is set to 0.8; and we set the default unconditional training probability $p_{\text{uncond}}$ to 0.1.
For the diffusion model, we {follow the process in Tashiro et al. (2021) \cite{CSDI}\footnote{{Details of the diffusion step encoding are presented in Appendix \ref{app:Archi details}.}} and} implement a quadratic noise schedule with an initial noise level of 0.0001.
Time steps are set to follow quadratic distribution sampling combined with the denoising diffusion implicit models (DDIM) \cite{ddim} approach. 
In training, except for that in {BERT-base}, all model parameters are updated through the Adam optimizer.

For evaluation, we follow conventional TSF studies \cite{patchTST, dlinear, CSDI} to employ mean squared error (MSE) and mean absolute error (MAE) as evaluation metrics, where lower scores mean better performance.\footnote{{We present the details of the evaluation metrics in Appendix \ref{app:mse and mae}.}}
We evaluate the models on the test data of all domains in the benchmark dataset, and report the average performance under short-, medium-, and long-term prediction lengths (e.g., 6, 12, and 18 months for monthly frequency data) correlated with the frequency of each domain.
We experiment with various hyperparameter combinations\footnote{Detailed parameter configurations are provided in Appendix \ref{app:Hyperparameters of MCD-TSF}.} and select the optimal configuration based on the performance on the validation set. 
To ensure statistical reliability, all experiments are repeated three times with different random seeds, and final results are the average of the three runs.

\begin{table}[t]
  \centering
  \caption{{The (a) MSE and (b) MAE of baselines and MCD-TSF (i.e., DIFF+TAA+TTF).
  The ``Avg.'' is the average performance on the test sets of eight domains.
  The first and second best results (the lower the better) on each domain are marked by \textbf{boldface} and \underline{underlines}, respectively.
  }}
  \label{tab:baselines}
  \vspace{-0.5em}
  \begin{subtable}[t]{\textwidth}
    \centering
    \caption{MSE Results}
    \label{tab:comp_baselines_mse}
    \begin{tabular}{l|cccccccc||c}
      \toprule
        & \textbf{Agr}   & \textbf{Cli}   & \textbf{Eco}   & \textbf{Ene}   & \textbf{Env}   & \textbf{Hea}   & \textbf{Soc}   & \textbf{Tra}   & \textbf{Avg.}\\
      \midrule
      DIFF           & 0.992        & 1.794 & 0.320 & 0.417 & 0.339 & 1.997 & 1.242 & 0.133 & 0.904\\
      DIFF+TAA       & 0.540        & 1.717 & 0.284 & 0.401 & 0.284 & 1.867 & 1.161 & 0.102 & 0.794\\
      DIFF+TAA-T     & 0.270        & \underline{1.654} & \underline{0.281} & \underline{0.241} & \underline{0.281} & 1.860 & \underline{1.126} & 0.101 & 0.726\\
      DIFF+TTF       & 0.341        & 1.813 & 0.411 & 0.318 & 0.334 & 1.837 & 1.222 & 0.097 & 0.796\\
      DIFF+TTF-T     & \underline{0.259}        & 1.667 & 0.311 & 0.305 & 0.283 & \underline{1.700} & 1.147 & \underline{0.096} & \underline{0.721}\\
      MCD-TSF
      & \textbf{0.222} & \textbf{1.583} & \textbf{0.249} & \textbf{0.153} & \textbf{0.275} & \textbf{1.496} & \textbf{1.035} & \textbf{0.093} & \textbf{0.638}\\
      \bottomrule
    \end{tabular}
  \end{subtable}
  \begin{subtable}[t]{\textwidth}
    \centering
    \caption{MAE Results}
    \label{tab:comp_baselines_mae}
    \begin{tabular}{l|cccccccc||c}
      \toprule
    & \textbf{Agr}   & \textbf{Cli}   & \textbf{Eco}   & \textbf{Ene}   & \textbf{Env}   & \textbf{Hea}   & \textbf{Soc}   & \textbf{Tra}  & \textbf{Avg.} \\
      \midrule
      DIFF           & 0.717 & 1.325 & 0.814 & 0.365 & 0.424 & 0.980 & 0.678 & 0.213 & 0.689\\
      DIFF+TAA       & 0.486 & 1.301 & 0.638 & 0.348 & 0.382 & 0.954 & \underline{0.624} & 0.174 & 0.613\\
      DIFF+TAA-T     & \underline{0.356} & \underline{1.156} & \underline{0.423} & \underline{0.315} & \underline{0.380} & 0.994 & 0.646 & \underline{0.171} & \underline{0.554}\\
      DIFF+TTF       & 0.552 & 1.504 & 0.448 & 0.392 & 0.419 & 0.944 & 1.127 & 0.178 & 0.695\\
      DIFF+TTF-T     & 0.381 & 1.237 & 0.479 & 0.350 & 0.382 & \underline{0.864} & 0.654 & 0.173 & 0.564\\
      MCD-TSF
      & \textbf{0.322} & \textbf{0.971} & \textbf{0.374} & \textbf{0.293} & \textbf{0.379} & \textbf{0.811} & \textbf{0.569} & \textbf{0.159} & \textbf{0.484}\\
      \bottomrule
    \end{tabular}
  \end{subtable}
  \vspace{-0.5em}
\end{table}

\section{Results and Analyses}
\subsection{Overall Results}

\begin{table}[tp]
  \centering
  \caption{Performance (MSE and MAE) comparison between our approach and existing studies, which are categorized into several groups based on their model types.
  }
  \label{tab:sota}
  \vspace{-0.5em}
  \begin{subtable}[t]{\textwidth}
    \centering
    \caption{MSE Results}
    \label{tab:comp_baselines_mse}
\begin{tabular}{l|c|c|c|c|c|c|c|c||c}
\toprule[2pt]
    &      \textbf{Agr} & \textbf{Cli} & \textbf{Eco} & \textbf{Ene} & \textbf{Env} & \textbf{Hea} & \textbf{Soc} & \textbf{Tra} & \textbf{Avg.} \\
\midrule
\multicolumn{1}{l|}{Reformer}         & 2.228       & 1.334   & 0.805   & 0.491  & 0.320       & 1.730  & 1.410       & 0.346   & 1.083        \\ 
\multicolumn{1}{l|}{Autoformer}       & 0.504       & 1.948   & 0.305   & 0.413  & 0.377       & \underline{1.434}  & 1.287       & 0.159   & 0.803        \\ 
\multicolumn{1}{l|}{FEDformer}        & 0.603       & 2.396   & 0.290   & 0.410  & 0.303       & 1.598  & 1.526       & 0.159   & 0.910        \\
\multicolumn{1}{l|}{PatchTST}        & 0.386       & 1.580   & 0.293   & 0.220  & \textbf{0.254}       & 1.557  & 1.066       & 0.124   & 0.685        \\ 
\multicolumn{1}{l|}{{HCAN}}             & 0.460       & 1.253   & 0.379   & 0.464  & 0.287       & 1.883  & \underline{1.038}       & 0.254   & 0.752        \\ 
\midrule
\multicolumn{1}{l|}{FiLM}             & 0.675       & 2.728   & 0.437   & 0.354  & 0.285       & 2.009  & 1.834       & 0.158   & 1.060        \\ 
\multicolumn{1}{l|}{DLinear}          & 1.581       & \textbf{0.950}   & 1.278   & 0.311  & 0.283       & 1.554  & 1.415       & 0.281   & 0.956        \\ 
\multicolumn{1}{l|}{TimeMixer++}      & 0.336       & 1.380   & 0.330   & 0.229  & 0.290       & 1.866  & 1.173       & \underline{0.101}   & 0.713        \\ 
\midrule
\multicolumn{1}{l|}{Timemachine}      & \underline{0.304}       & 1.797   & 0.281   & 0.264  & 0.276       & 1.666  & 1.341       & 0.102   & 0.753        \\ 
\midrule
\multicolumn{1}{l|}{CSDI}             & 2.096       & 1.264   & 1.492   & 0.410  & 0.291       & 1.685  & 1.112       & 0.115   & 1.058        \\ 
\multicolumn{1}{l|}{D3VAE}            & 1.166       & \underline{1.220}   & 0.661   & 0.286  & 0.755       & 1.815  & 1.098       & 0.232   & 0.904        \\
\midrule
\multicolumn{1}{l|}{{FPT}}             & 0.448       & 1.775   & 0.303   & 0.234  & 0.283       & 1.551  & 1.047       & 0.182   & 0.727        \\ 
\midrule
\multicolumn{1}{l|}{TimeLLM}          & 0.425       & 1.417   & \underline{0.276}   & \underline{0.203}  & 0.277       & 1.519  & 1.155       & 0.141   & \underline{0.676}        \\ 
\multicolumn{1}{l|}{MM-TSF}           & 0.714       & 1.278   & 1.103   & 0.610  & 0.336       & \textbf{1.368}  & 1.067       & 0.225   & 0.837        \\ 
\midrule
\multicolumn{1}{l|}{GLAFF}            & 0.432       & 2.021   & 0.391   & 0.217  & 0.283       & 1.563  & 1.039       & 0.112   & 0.757        \\ 
\multicolumn{1}{l|}{TimeLinear}       & 0.497       & 1.723   & 0.327   & 0.261  & 0.284       & 1.556  & 1.518       & 0.163   & 0.791        \\ 
\midrule
\midrule
\multicolumn{1}{l|}{MCD-TSF (ours)}    & \textbf{0.222} & 1.583   & \textbf{0.249} & \textbf{0.153} & \underline{0.275}       & 1.496  & \textbf{1.035}       & \textbf{0.093} & \textbf{0.638}        \\
\bottomrule[2pt]
\end{tabular}%
  \end{subtable}
  \begin{subtable}[t]{\textwidth}
    \centering
    \caption{MAE Results}
    \label{tab:comp_baselines_mae}
\begin{tabular}{l|c|c|c|c|c|c|c|c||c}
\toprule[2pt]
    &     \textbf{Agr} & \textbf{Cli} & \textbf{Eco} & \textbf{Ene} & \textbf{Env} & \textbf{Hea} & \textbf{Soc} & \textbf{Tra} & \textbf{Avg.} \\
\midrule
\multicolumn{1}{l|}{Reformer}        & 1.150       & 1.094   & 0.745   & 0.531  & 0.448       & 0.860  & 0.848       & 0.499   & 0.771        \\ 
\multicolumn{1}{l|}{Autoformer}       & 0.538       & 1.394   & 0.439   & 0.454  & 0.438       & 0.895  & 0.725       & 0.270   & 0.644        \\ 
\multicolumn{1}{l|}{FEDformer}        & 0.628       & 1.674   & 0.424   & 0.458  & 0.412       & 0.943  & 0.926       & 0.291   & 0.719        \\
\multicolumn{1}{l|}{PatchTST}        & 0.404       & 1.240   & 0.408   & \underline{0.307}  & \textbf{0.367}       & 0.863  & 0.576       & 0.229   & 0.549        \\ 
\multicolumn{1}{l|}{HCAN}            & 0.552       & 0.955   & 0.465   & 0.452  & 0.420       & 0.869  & 0.587       & 0.402   & 0.587        \\ 
\midrule
\multicolumn{1}{l|}{FiLM}             & 0.577       & 1.870   & 0.540   & 0.426  & 0.388       & 1.143  & 1.046       & 0.292   & 0.785        \\ 
\multicolumn{1}{l|}{DLinear}          & 0.997       & \textbf{0.840}   & 0.932   & 0.380  & 0.382       & 0.878  & 0.848       & 0.401   & 0.707        \\ 
\multicolumn{1}{l|}{TimeMixer++}      & 0.400       & \underline{0.909}   & 0.430   & 0.337  & 0.448       & 0.933  & 0.589       & \underline{0.169}   & \underline{0.526}       \\ 
\midrule
\multicolumn{1}{l|}{Timemachine}     & \underline{0.358}       & 1.046   & 0.408   & 0.373  & 0.381       & 0.878  & 0.606       & 0.174   & 0.528        \\ 
\midrule
\multicolumn{1}{l|}{CSDI}            & 1.048       & 1.028   & 0.968   & 0.433  & 0.391       & 0.857  & \underline{0.573}       & 0.238   & 0.692        \\ 

\multicolumn{1}{l|}{D3VAE}            & 0.822       & 1.024   & 0.612   & 0.387  & 0.730       & 0.893  & 0.632       & 0.367   & 0.683        \\ 
\midrule
\multicolumn{1}{l|}{FPT}              & 0.510       & 1.057   & 0.437   & 0.362  & 0.406       & 0.844  & 0.587       & 0.317   & 0.565        \\ 
\midrule
\multicolumn{1}{l|}{TimeLLM}          & 0.491       & 0.921   & \underline{0.397}   & 0.336  & 0.398       & 0.818  & 0.631       & 0.227   & 0.527        \\ 
\multicolumn{1}{l|}{MM-TSF}          & 0.607       & 1.041   & 0.820   & 0.601  & 0.467       & \textbf{0.760}  & 0.598       & 0.402   & 0.662        \\ 
\midrule
\multicolumn{1}{l|}{GLAFF}            & 0.553       & 1.208   & 0.499   & 0.326  & 0.398       & 0.848  & 0.576       & 0.206   & 0.576        \\ 
\multicolumn{1}{l|}{TimeLinear}       & 0.549       & 1.034   & 0.457   & 0.389  & 0.387       & 0.851  & 0.884       & 0.300   & 0.606        \\ 
\midrule
\midrule
\multicolumn{1}{l|}{MCD-TSF (ours)}    & \textbf{0.322} & 0.971   & \textbf{0.374} & \textbf{0.293} & \underline{0.379} & \underline{0.811} & \textbf{0.569} & \textbf{0.159} & \textbf{0.484}      \\
\bottomrule[2pt]
\end{tabular}
\end{subtable}
\vspace{-0.5em}
\end{table}

The comparative results between our approach and different baseline models are shown in Table \ref{tab:baselines}, where the average performance of each model on all datasets is also reported.
There are several observations.
First, the ``DIFF+TAA'' and ``DIFF+TTF'' consistently outperform the standalone ``DIFF'' model.
This improvement is attributed to the enhanced guidance provided by additional timestamps or text inputs during the denoising process, highlighting the importance of multimodal enhancement to conventional TSF modeling.
Second, {``DIFF+TAA-T'' and ``DIFF+TTF-T'' that utilize both timestamps and text to enhance time series achieve better performance than ``DIFF+TAA'' and ``DIFF+TTF'' that only use one of timestamps and text to enhance time series.
The results demonstrate} that the synergistic fusion of timestamps and textual information yields superior help compared to using either modality independently.
Third, the proposed {MCD-TSF} (equivalent to "DIFF+TAA+TTF")
consistently outperforms other baselines (i.e., the ``DIFF+TAA-T'' and ``DIFF+TTF-T'').
This advantage stems from two complementary mechanisms: the TAA module enhances temporal relationship modeling {across the entire input time series} through timestamp features, while the TTF module performs finer-grained text fusion and controls the strength of text guidance.
{
The two modules enable MCD-TSF to effectively integrate coarse-grained and fine-grained temporal patterns into the time series, resulting in more accurate and robust forecasts across diverse domains.
}

We further compare our approach with existing studies and present the results\footnote{{For each domain, Table \ref{tab:sota} reports the average model performance under different configurations specified in Section \ref{sec:implementation}, where the full result on each configuration is reported in Appendix \ref{app:full results}.}} in Table \ref{tab:sota}, with several observations drawn from it.
First, overall, MCD-TSF outperforms all existing models across six model-type categories, which clearly demonstrates its versatile effectiveness.
Second, when particularly compared with probabilistic models {(i.e., CSDI, D3VAE}), MCD-TSF shows lower {MSE and MAE on seven out of eight domains}. 
This advantage favors from MCD-TSF's effective integration of multimodal conditioning, which provides stronger supervision during the denoising process. 
Third, compared with existing multimodal approaches (including those models that also incorporate textual information or timestamp features as we do), MCD-TSF achieves better performance in {on six out of eight domains}.
{This improvement is attributed to that separately encoding timestamps and text and then progressively integrating them with the time series allows the MCD-TSF to fully leverage both modalities to enrich temporal representations and boost forecasting accuracy.}
In addition, MCD-TSF performs a controlled text fusion, effectively mitigating the influence of textual noise, so that further improves its TSF ability.

\subsection{Effect of Timestamp Condition}
\label{sec:timestamp-effect}

{
To explore the effect of the timestamp, we run different configurations of the timestamp weight $\lambda$ to control the contribution of timestamp information to enhancing the time series representation, and perform a case study to visualize the effect of timestamps on model performance.
}

{
For the effect of $\lambda$, we try different values, including 0.2, 0.4, 0.6, 0.8, and 1.0, and report the results on \textit{agriculture}, \textit{economy}, \textit{social good}, and \textit{traffic} domains in Figure \ref{fig:timestamp-weight}.\footnote{{The results on the other four domains are reported in Appendix \ref{app:timestamp-weight}.}}
Herein, higher $\lambda$ refers to a higher contribution of the timestamp in the fusion process of the time series.
The results show that our approach achieves better performance across all domains with higher values of $\lambda$, which confirms the effectiveness of leveraging timestamp information for TSF.
}

\begin{figure}[t]
\centering
    \includegraphics[width=\linewidth]{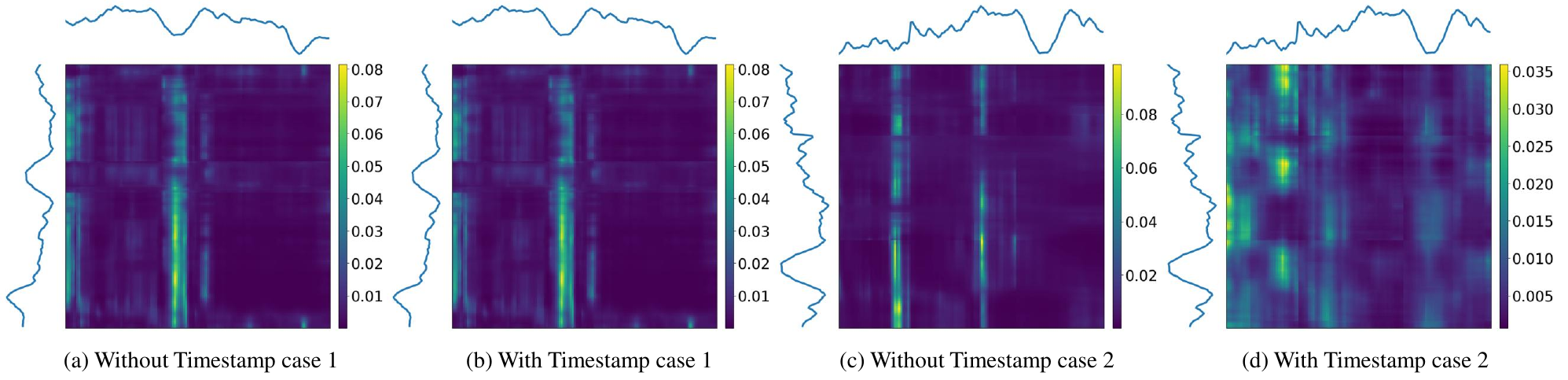}
    \caption{{Visual of the attention of our approach across data points of two cases with different settings, where darker colors refer to higher attention weights. 
    Figures (a) and (c) are the attentions from models without using timestamps for case 1 and case 2, respectively; and Figures (b) and (d) are the attentions from models with timestamps for  case 1 and case 2, respectively.}}
\label{fig:timestamp heatmap}
\end{figure}

{
To further analyze the importance of timestamp fusion for time series modeling, we present the attention heatmaps of two cases from the fourth layer of the TAA module during MCD-TSF's prediction process with and without the timestamp input on the \textit{Energy} domain dataset in Figure \ref{fig:timestamp heatmap}, where brighter colors stand for higher attention weights.
There are following observations.
First, we observe that both models with and without timestamps present brighter colors at the peaks and troughs of the time series, indicating their capability to focus on numerically critical segments (i.e., local maxima and minima). 
Second, the timestamp-enhanced model displays equally bright colors in additional segments of the time series, accompanied by intricate attention patterns. 
This demonstrates that the TAA module is able to effectively enhance the model's attention to temporally relevant features and thus improve the representations of the time series for TSF.
}

\begin{figure}[t]
    \centering
    \includegraphics[width=\linewidth]{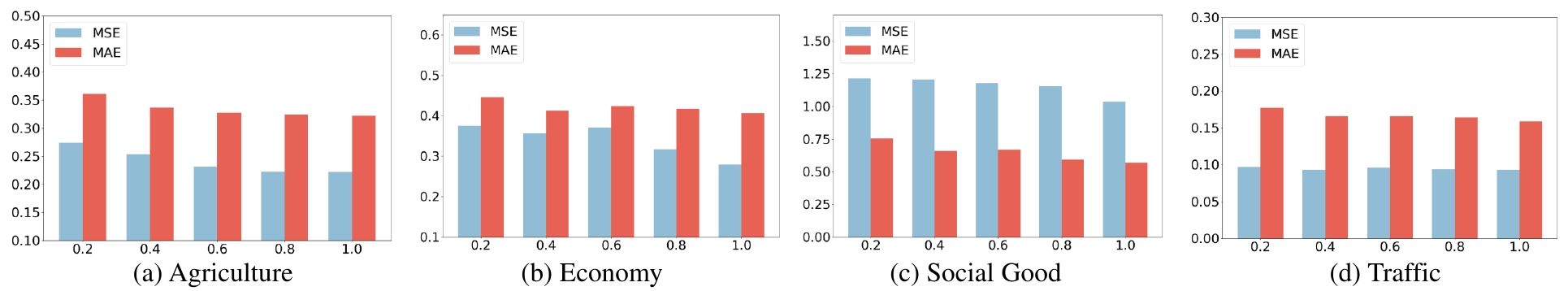}
    \caption{{Results of our approach on different domains when configured with various values of the timestamp weights (i.e., the $\lambda=0.2, 0.4, 0.6, 0.8, 1.0$).}}
    \label{fig:timestamp-weight}
\end{figure}

\subsection{Effect of Text Condition}
\label{sec:text-effect}

\begin{figure}[t]
\centering
    \includegraphics[width=\linewidth]{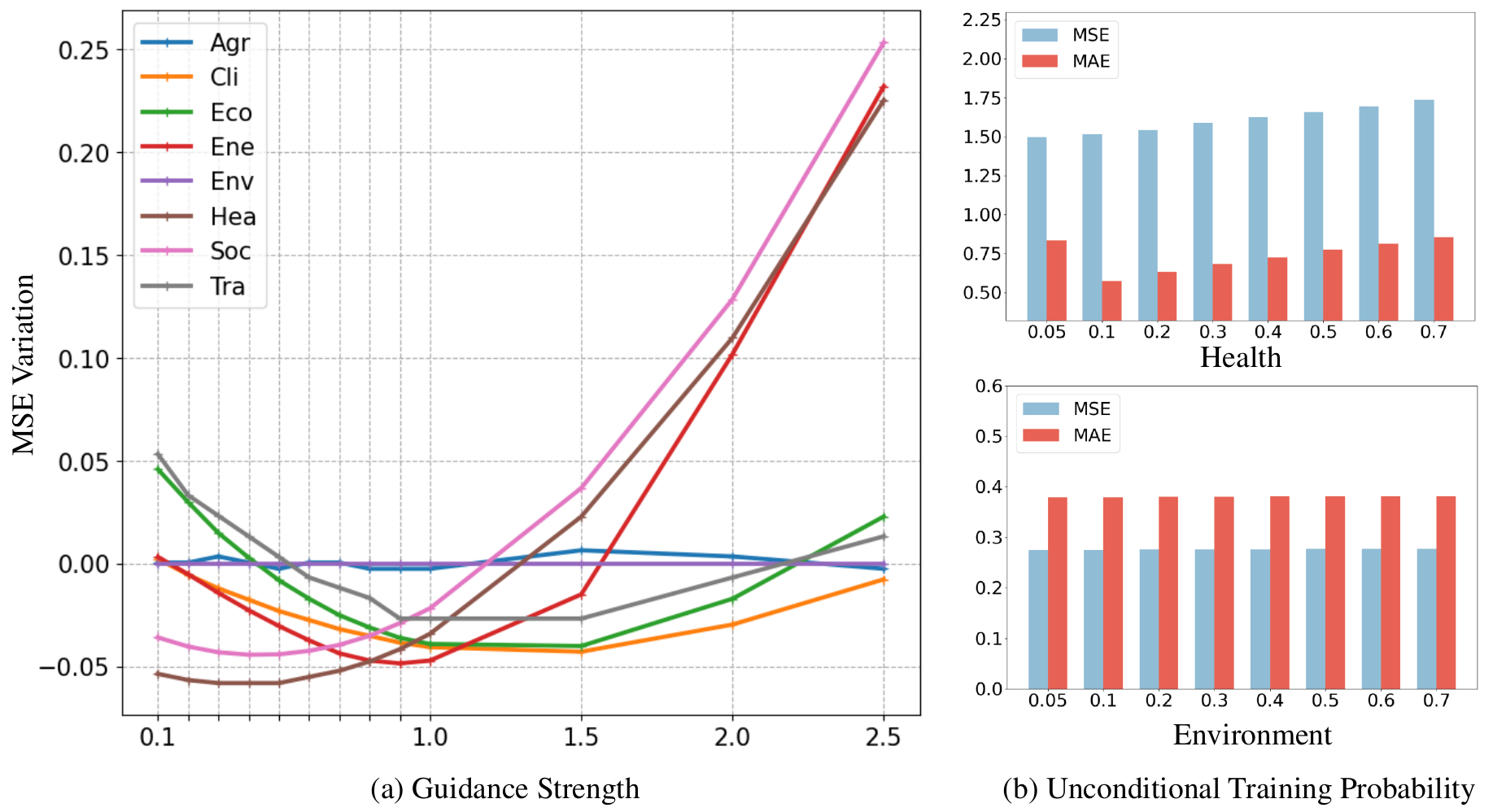}
    \caption{
    {
    Figure (a) illustrates how the model's MSE varies with different guidance strengths across datasets from multiple domains, where the reported MSE for each configuration is calculated by subtracting the mean MSE across all configurations within the respective domain.
    Figure (b) displays the model performance with changes in the unconditional training probability.}
    }
\label{fig:guidance strength}
\end{figure}

{
To investigate the effect of text modality on TSF, we analyze the effect of two hyperparameters, namely, the textual guidance strength \( w \) used only in inference to control the contribution of text to TSF, and the unconditional training probability \( p_{\text{uncond}} \) employed only in training to influence how much text information is learned to guide TSF.
The following are the details.
}

{
For textual guidance strength \( w \), we adjust its value within the range $[0.1, 2.5]$ (specifically, we try the values of 0.1, 0.2, 0.3, 0.4, 0.5, 0.6, 0.7, 0.8, 0.9, 1.0, 1.5, 2.0, and 2.5), and analyze performance variations across different domains.
We present the curve of MSE variation \footnote{{The MSE variation for a domain is calculated by subtracting the domain’s mean MSE from the obtained MSEs under different guidance strength $w$, thereby enabling a more intuitive comparison of the trends in model performance across different domains.}} on the test sets of eight domains with respect to the value of $w$ in Figure \ref{fig:guidance strength}(a).
There are following observations.
First, overall, on most domains (e.g., the \textit{economy}), the MSE initially decreases and then subsequently increases as the value of guidance strength grows, indicating diminished model performance under both insufficient and excessive textual guidance. 
The following are the explanations.
An insufficient \( w \) fails to effectively leverage textual information for prediction guidance, while an overly large \( w \) over-prioritizes textual inputs at the expense of historical time series patterns and temporal information.
Second, there are domains where the fluctuation of the curve is not significant (e.g., the \textit{environment} domain), which is potentially due to the extreme sparsity of textual data in the domain data.
Third, most domains present relatively flat error curves near their minima, indicating the robustness of the proposed model in leveraging text information to enhance the representation of time series. 
}

{
For unconditional training probability \( p_{\text{uncond}} \),
we try the following values: 0.05, 0.1, 0.2, 0.3, 0.4, 0.5, 0.6, and 0.7.
Figure \ref{fig:guidance strength}(b) presents two representative results\footnote{{The results on the other domains are reported in Appendix \ref{app:Unconditional-Training-Probability}.}} from \textit{health} and \textit{environment} domains.
For the results on the \textit{health} domain, the MSE and MAE increase with higher $p_{\text{uncond}}$.
This observation is intuitive since a higher $p_{\text{uncond}}$ indicates more textual information is masked during training, which prevents the model from effectively leveraging the textual information to enhance the time series representations and thus leads to inferior performance.
For the results on the \textit{environment} domain, the change in model performance is not noticeable, which aligns with the results presented in \ref{fig:guidance strength}(a) and is attributed to the extreme sparsity of textual data in this domain.
}

\subsection{Case Study}

\begin{figure}[t]
\centering
    \includegraphics[width=\linewidth]{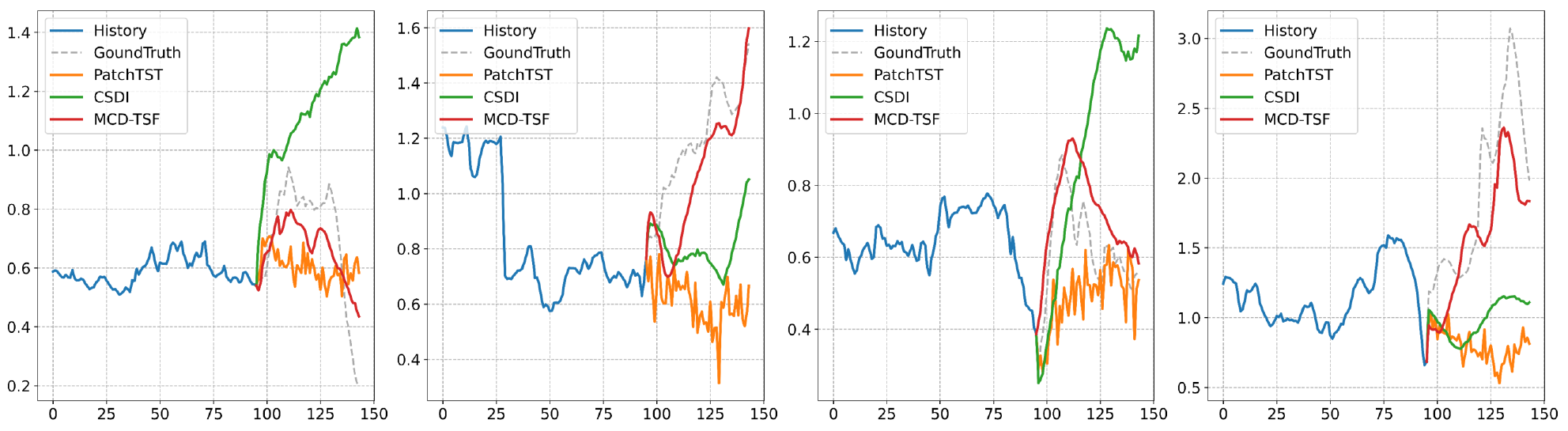}
    \caption{{
    Case study with four examples from \textit{energy} domains.
    In all cases, the historical time series and the gold standard future time series are marked by blue and dashed grey lines, respectively.
    The predictions from PatchTST, CSDI, and our approach (i.e., MCD-TSF) are represented by orange, green, and red lines, respectively.
    }}
    \label{fig:case_study}
\end{figure}

To qualitatively evaluate the advantages of TSF with multimodal input, Figure \ref{fig:case_study} presents several prediction results from {two baselines (i.e., PatchTST \cite{patchTST}, and CSDI \cite{CSDI})} and MCD-TSF in the \textit{energy} domain from the experiment dataset.
The selected cases illustrate non-stationary time series where accurate forecasting requires models to comprehensively understand dynamic variations rather than merely replicating historical patterns. 
Observations reveal that all baselines and our approach produce prediction curves closely aligned with the ground truth for short-term forecasting.
However, in extending forecasting horizons, the curves from both PatchTST and CSDI show significant deviations from the ground truth.
Particularly, CSDI captures general trend similarities with the ground truth but fails to predict inflections in the forecast horizon,
which suggests that although, to some extent, CSDI learns {the relations among different timestamps}, it still lacks robustness for longer-term predictions. 
{
PatchTST predicts variations in the time series that closely resemble the patterns in historical time series, highlighting that the model heavily relies on past data.
In contrast, MCD-TSF accurately captures the complete temporal evolution throughout the forecast period, maintaining close alignment with the Ground Truth. 
This indicates that MCD-TSF effectively learns the intrinsic dynamics of time series through multimodal assistance of temporal and textual information.
}

\section{Conclusion}
In this paper, we propose a novel multimodal diffusion model for TSF, named MCD-TSF, which incorporates extra two modalities, e.g., timestamps and text, to model the temporal relationships among data points and enrich the semantic information of the time series.
Specifically,
upon a Transformer-based diffusion implementation,
a multimodal fusion module with TAA and TTF is proposed to incorporate 
the aforementioned multimodal inputs,
thereby progressively enhancing the representations of time series data.
Classifier-free guidance (CFG) is utilized to further dynamically control the impact of the text information on the enhanced time series representations.
Extensive experiment results and analyses on real-world datasets across eight domains show that MCD-TSF achieves better predictions compared to strong baselines and existing studies for TSF.

\bibliographystyle{plain}
\bibliography{my_paper}


\newpage

\appendix

\section{Timestamp Encoding}
\label{app:encoding of timestamp}
{
The dataset utilized in this study incorporates three sampling frequencies: daily, weekly, and monthly. 
In the case of the daily dataset, it is represented by three-dimensional features: [$\texttt{DayOfWeek}$, $\texttt{DayOfMonth}$, $\texttt{DayOfYear}$]. 
For the weekly dataset, two-dimensional features are employed: [$\text{DayOfMonth}$, $\texttt{WeekOfYear}$]. 
As for the monthly dataset, the feature of [$\texttt{MonthOfYear}$] is utilized. 
The different time encoding calculation formulas are as follows:
\begin{equation}
    \begin{aligned}
\texttt{DayOfWeek} & = \frac{i_{d}}{6}-0.5\\
\texttt{DayOfMonth} & = \frac{{i_{d}-1}}{30}-0.5\\
\texttt{DayOfYear} & = \frac{{i_{d}-1}}{365}-0.5\\
\texttt{WeekOfYear} & = \frac{{i_{w}-1}}{52}-0.5\\
\texttt{MonthOfYear} & = \frac{i_{m}}{11}-0.5,\\
\end{aligned}
\end{equation}
where $i_d$, $i_w$, and $i_m$ represent the daily, weekly, and monthly indexes, respectively.
}

\section{Text Input Design}
\label{app:text input}
{
In our experiment, we collect reports of all dates within the historical period to compile a text document that represents textual data describing the time series.
Below, we present our text input template (the placeholders are marked by angle brackets) and an example of it.
}

\begin{tcolorbox}[title=Template, mybox]
Below is historical reporting information over the past 
\textbf{〈history\_length〉} \textbf{〈sampling\_period〉}s concerning the 
\textbf{〈target\_variable〉}. Based on these reports, predict the potential 
trends and anomalies of the \textbf{〈target\_variable〉} for the next 
\textbf{〈prediction\_length〉} \textbf{〈sampling\_period〉}s.

\smallskip
\noindent
\textbf{〈start\_date1〉} to \textbf{〈end\_date1〉}: \textbf{〈report1〉}.  
\textbf{〈start\_date2〉} to \textbf{〈end\_date2〉}: \textbf{〈report2〉} …  
\end{tcolorbox}

\vspace{1em}

\begin{tcolorbox}[title=Example, mybox]
Below is historical reporting information over the past 
\textbf{14} \textbf{days} concerning the \textbf{stock price}. Based on these 
reports, predict the potential trends and anomalies of the stock price for the 
next \textbf{7} \textbf{days}.

\smallskip
\noindent
\textbf{2025-04-01} to \textbf{2025-04-07}: Company A’s shares rose steadily due 
to strong quarterly earnings.  
\textbf{2025-04-08} to \textbf{2025-04-14}: Market volatility increased amid 
macroeconomic uncertainty. …
\end{tcolorbox}

\section{Diffusion Step Encoding}
\label{app:Archi details}

For the $m$-th data point in the historical window, we use a 128-dimensional temporal embedding:
\begin{equation}
    s_{\text {embedding }}\left({m}\right)=\left(\sin \left({m} / \tau^{0 / 64}\right), \ldots, \sin \left({m} / \tau^{63 / 64}\right), \cos \left({m} / \tau^{0 / 64}\right), \ldots, \cos \left({m} / \tau^{63 / 64}\right)\right)
\end{equation}
{
where $\tau = 10000$. Note that although our proposed model already includes global temporal information provided by timestamps, we still retain the use of temporal embedding introduced by Tashiro et al. (2021) \cite{CSDI} because it focuses more on time encoding starting from the historical window.
}

\section{Evaluation Metrics}
\label{app:mse and mae}

{We compute the MSE and MAE by
\begin{equation}
    \begin{aligned}
        M A E&=\operatorname{mean}\left(\left|\widehat{\mathcal{Y}}-\mathcal{Y}^*\right|\right) \\
M S E&=\sqrt{\operatorname{mean}\left(\left|\widehat{\mathcal{Y}}-\mathcal{Y}^*\right|\right)}.
    \end{aligned}
\end{equation}
where $\widehat{\mathcal{Y}}$ and $\mathcal{Y}^*$ stand for the model predictions and the gold standard, respectively.
}

\section{Hyperparameters of MCD-TSF}
\label{app:Hyperparameters of MCD-TSF}

{
As illustrated in Appendix \ref{app:encoding of timestamp}, the dimension of timestamp features is related to their recording frequency, with a daily frequency having a dimension of 3, a weekly frequency having a dimension of 2, and a monthly frequency having a dimension of 1. 
Besides, the dimension of diffusion step embeddings is 128 and the number of attention heads is 8.
Table \ref{tab: Hyperparameters of MCD-TSF} summarizes these hyperparameters.
}
\begin{table}[t]
\centering
\caption{Hyperparameters of MCD-TSF}
\label{tab: Hyperparameters of MCD-TSF}
\begin{tabular}{l|ccc}
\toprule[2pt]
         & \textbf{Monthly} & \textbf{Weekly} & \textbf{Daily} \\
\midrule
Attention heads                & 8                & 8               & 8              \\
Timestamp features dim.        & 1                & 2               & 3              \\
diffusion step embeddings dim. & 128              & 128             & 128            \\
\bottomrule[2pt]
\end{tabular}
\end{table}

\section{Supplementary Results}
\label{Supplementary Results}

\subsection{Module Order Comparison}
\label{app:exp model order}

{
In experiments, we conduct a comparative analysis to investigate the impact of sequential arrangements between the TAA module and TTF module on model performance in multimodal fusion.
The results are reported in Table \ref{tab:module order}, where our approach (TAA-TTF) that firstly fuses the timestamp to the time series and then integrates the text into the time series achieves better performance than the TTF-TAA model that performs the fusion operations in the reverse order.
}
\begin{table}[t]
\centering
\caption{The results of models with different orders to incorporate timestamps and text information into the time series representations through TAA and TTF.}
\label{tab:module order}
\begin{tabular}{c|cc|cc|cc|cc}
\toprule[2pt]
\multirow{2}{*}{} & \multicolumn{2}{c|}{Agr} & \multicolumn{2}{c|}{Ene} & \multicolumn{2}{c|}{Eco} & \multicolumn{2}{c}{Tra} \\ 
\midrule
                  & MSE             & MAE            & MSE             & MAE            & MSE             & MAE            & MSE              & MAE                      \\
\midrule
TTF-TAA           & 0.247           & 0.328          & 0.215           & 0.354          & 0.282               & 0.383            & 0.102           & 0.175          \\
TAA-TTF           & \textbf{0.222}           & \textbf{0.322}          & \textbf{0.153}           & \textbf{0.293}          & \textbf{0.249}                   & \textbf{0.374}            & \textbf{0.093}           & \textbf{0.159}       \\
\bottomrule[2pt]
\end{tabular}%
\end{table}

\subsection{Results of Different Horizons Setting}
\label{app:full results}

{
In Table \ref{tab:comp_baselines_all_horizons_mse1} and Table \ref{tab:comp_baselines_all_horizons_mse2}, we present the full performance comparison (i.e., MSE and MAE on the eight domains) between the MCD-TSF model and existing studies with varying prediction horizons, whose settings are specified in Section \ref{sec:implementation}.
}
\begin{table}[t]
  \centering
  \caption{Comparative MSE results between our approach and baseline models in all horizons.}
  \begin{subtable}[t]{\textwidth}
    \centering
    \caption{MSE Results 1}
    \label{tab:comp_baselines_mse}
\resizebox{\columnwidth}{!}{
\begin{tabular}{l|ccc|ccc|ccc|ccc}
\toprule[2pt]
            & \multicolumn{3}{c|}{Agriculture} & \multicolumn{3}{c|}{Climate} & \multicolumn{3}{c|}{Economy} & \multicolumn{3}{c}{Energy} \\
            \midrule
            & 6      & 12    & 18    & 12    & 24    & 48    & 6     & 12    & 18    & 12    & 24    & 48   \\
\midrule
Reformer    & 1.694  & 2.695 & 2.295 & 1.236 & 1.305 & \underline{1.461} & 0.638 & 0.792 & 0.986 & 0.426 & 0.468 & 0.580 \\
Autoformer  & 0.318  & 0.484 & 0.710 & 1.266 & 1.902 & 2.676 & 0.267 & 0.300 & 0.348 & 0.402 & 0.403 & 0.438 \\
FEDformer   & 0.431  & 0.631 & 0.747 & 1.762 & 2.365 & 3.062 & 0.250 & 0.303 & \textbf{0.317} & 0.317 & 0.406 & 0.509 \\
PatchTST    & 0.158  & 0.400 & 0.598 & \textbf{0.568} & 1.174 & 2.998 & 0.224 & 0.282 & 0.373 & \underline{0.093} & 0.202 & 0.365 \\
HCAN        & 0.363  & 0.549 & \underline{0.467} & 0.887 & \underline{1.129} & 1.742 & 0.328 & 0.367 & 0.443 & 0.115 & \underline{0.162} & 1.115 \\
\midrule
FiLM        & 0.432  & 0.556 & 1.038 & 2.360 & 2.548 & 3.276 & 0.356 & 0.386 & 0.570 & 0.286 & 0.354 & 0.422 \\
DLinear     & 0.985  & 1.015 & 2.745 & 0.841 & 0.887 & \textbf{1.124} & 0.816 & 1.252 & 1.766 & 0.244 & 0.299 & 0.392 \\
TimeMixer++ & 0.151  & 0.354 & 0.502 & 0.635 & 1.088 & 2.418 & \textbf{0.181} & \textbf{0.216} & 0.594 & \textbf{0.088} & 0.220 & 0.379 \\
\midrule
Timemachine & 0.140   & \underline{0.285} & 0.486 & 0.831 & 1.515 & 3.046 & \underline{0.182} & 0.273 & 0.389 & 0.122 & 0.295& 0.376 \\
\midrule
CSDI        & 1.995  & 2.159 & 2.136 & 0.762 & 1.311 & 1.721 & 0.970  & 1.590 & 1.918 & 0.191 & 0.491 & 0.547 \\
D3VAE        & \textbf{0.097} & 1.143 & 2.259 & 0.718 & \textbf{1.124} & 1.819 & 0.464 & 0.628 & 0.891 & 0.181 & 0.240 & 0.439 \\
\midrule
FPT         & 0.299  & 0.450  & 0.596 & 0.631 & 1.840 & 2.853 & 0.267 & 0.312 & \underline{0.331} & 0.123 & 0.234 & 0.344 \\
\midrule
TimeLLM     & 0.312  & 0.480 & 0.483 & 0.830  & 1.270 & 2.151 & 0.219 & 0.258 & 0.350  & 0.103 & 0.197& \underline{0.309} \\
MM-TSF      & 1.333  & 1.395 & 1.415 & \underline{0.592} & 1.285 & 1.959 & 0.839 & 0.965 & 1.505 & 0.317 & 0.569 & 0.944 \\
\midrule
GLAFF       & 0.389  & 0.433 & 0.474 & 1.684 & 1.934 & 2.446 & 0.197 & 0.490  & 0.486 & 0.098 & 0.191 & 0.362 \\
TimeLinear  & 0.410   & 0.528 & 0.554 & 1.059 & 1.662 & 2.447 & 0.258 & 0.303 & 0.421 & 0.175 & 0.247 & 0.361 \\
\midrule 
\midrule
MCD-TSF     & \underline{0.110}   & \textbf{0.219} & \textbf{0.337} & 0.864 & 1.749 & 2.136 & 0.186 & \underline{0.245} & \textbf{0.317} & \textbf{0.088} & \textbf{0.153} & \textbf{0.219} \\
\bottomrule[2pt]
\end{tabular}
}
  \end{subtable}
  \vspace{0.1em}
  \begin{subtable}[ht]{\textwidth}
    \centering
    \caption{MSE Results 2}
    \label{tab:comp_baselines_all_horizons_mse2}
\resizebox{\columnwidth}{!}{
\begin{tabular}{l|ccc|ccc|ccc|ccc}
\toprule[2pt]
            & \multicolumn{3}{c|}{Environment} & \multicolumn{3}{c|}{Health} & \multicolumn{3}{c|}{Social Good} & \multicolumn{3}{c}{Traffic} \\
\midrule
            & 48     & 96    & 192   & 12    & 24    & 48   & 6      & 12    & 18    & 6    & 12    & 18   \\
\midrule
Reformer    & 0.308  & 0.316 & 0.336 & 1.555 & 1.798 & 1.839 & 1.287  & 1.443 & 1.502 & 0.267 & 0.343 & 0.429 \\
Autoformer  & 0.341  & 0.391 & 0.401 & 1.331 & 1.458 & \underline{1.513} & 0.940  & 1.452 & 1.469 & 0.141 & 0.159 & 0.178 \\
FEDformer   & 0.300  & 0.305 & 0.306 & 1.326 & 1.607 & 1.861 & 1.257  & 1.593 & 1.729 & 0.146 & 0.165 & 0.167 \\
PatchTST    & \textbf{0.251}  & \textbf{0.254} & \textbf{0.257} & 1.316 & 1.598 & 1.753 & 0.825  & 1.141 & 1.234 & 0.108 & 0.128 & 0.138 \\
HCAN        & 0.263  & 0.292 & 0.307 & 1.419 & 2.075 & 2.156 & 0.797  & 1.006 & 1.310 & 0.212 & 0.218 & 0.331 \\
\midrule
FiLM        & 0.304  & 0.282 & \underline{0.269} & 2.062 & 1.716 & 2.249 & 1.510  & 2.136 & 1.834 & 0.136 & 0.162 & 0.178 \\
DLinear     & 0.265  & 0.289 & 0.297 & 1.312 & 1.566 & 1.784 & 1.239  & 1.439 & 1.568 & 0.167 & 0.330 & 0.348 \\
TimeMixer++ & 0.294  & 0.279 & 0.297 & 1.392 & 1.729 & 2.477 & 0.925  & 1.102 & 1.492 & 0.086 & \underline{0.100} & 0.116 \\
\midrule
Timemachine & \underline{0.257}  & 0.288 & 0.282 & 1.367 & 1.581 & 2.050 & 1.012  & 1.478 & 1.532 & \underline{0.094} & 0.101 & \underline{0.110} \\
\midrule
CSDI        & 0.277  & 0.310 & 0.287 & 1.309 & 1.813 & 1.933 & 0.828  & 1.149 & 1.359 & 0.113 & 0.115 & 0.119 \\
D3VAE       & 0.344  & 0.443 & 0.486 & 2.230 & 1.742 & \textbf{1.473} & 0.936  & 1.155 & 1.204 & 0.213 & 0.228 & 0.255 \\
\midrule
FPT         & 0.276  & 0.285 & 0.287 & 1.432 & 1.499 & 1.722 & \underline{0.805}  & \underline{0.992} & 1.343 & 0.175 & 0.184 & 0.188 \\
\midrule
TimeLLM     & 0.269  & \underline{0.275} & 0.287 & 1.502 & \textbf{1.171} & 1.884 & 0.952  & 1.118 & 1.395 & 0.150 & 0.137 & 0.139\\
MM-TSF      & 0.320  & 0.332 & 0.356 & \underline{1.096} & \underline{1.438} & 1.570 & 0.827  & 1.166 & \underline{1.189} & 0.205 & 0.227 & 0.243  \\
\midrule
GLAFF       & 0.269  & 0.282 & 0.298 & 1.196 & 1.658 & 1.836 & \textbf{0.790} & 1.142 & \textbf{1.184} & 0.104 & 0.111 & 0.120 \\
TimeLinear  & 0.279 & 0.284 & 0.288  & 1.373 & 1.518 & 1.777 & 1.310  & 1.513 & 1.730 & 0.150 & 0.163 & 0.176 \\
\midrule
\midrule
MCD-TSF     & 0.258  & 0.276 & 0.290  & \textbf{1.026} & 1.566 & 1.895 & 0.814  & \textbf{0.950} & 1.342 & \textbf{0.085} & \textbf{0.097} & \textbf{0.098} \\
\bottomrule[2pt]
\end{tabular}
}
  \end{subtable}
\end{table}

\begin{table}[ht]
  \centering
  \caption{Comparative MAE results between our approach and baseline models in all horizons.}
 \label{tab:comp_baselines_all_horizons_mse2} 
  \begin{subtable}[t]{\textwidth}
    \centering
    \caption{MAE Results 1}
    \label{tab:comp_baselines_mse}
\resizebox{\columnwidth}{!}{
\begin{tabular}{c|ccc|ccc|ccc|ccc}
\toprule[2pt]
            & \multicolumn{3}{c|}{Agriculture} & \multicolumn{3}{c|}{Climate} & \multicolumn{3}{c|}{Economy} & \multicolumn{3}{c}{Energy} \\
\midrule
            & 6   & 12  & 18  & 12  & 24  & 48  & 6   & 12  & 18  & 12  & 24  & 48  \\
\midrule
Reformer    & 0.933 &1.209&1.309&0.927&0.964&\underline{1.023}&0.662&0.734&0.840&0.491&0.533&0.611\\
Autoformer  & 0.428 &0.538&0.648&0.892&1.075&1.343&0.414&0.438&0.467&0.471&0.479&0.501\\
FEDformer   & 0.533 &0.631&0.706&1.114&1.290&1.514&0.439&0.438&\textbf{0.395}&0.438&0.481&0.527\\
PatchTST    & 0.258 &0.426&0.528&0.594&\underline{0.817}&1.475&0.379&0.419&0.458&0.225&0.336&0.477\\
HCAN        & 0.511 &0.529&0.615&0.782&0.892&1.192&0.423&0.462&0.510&0.235&\textbf{0.287}&0.835\\
\midrule
FiLM        & 0.471 &0.559&0.702&1.224&1.243&1.639&0.479&0.509&0.632&0.425&0.451&0.499\\
DLinear     & 0.697 &1.015&1.279&0.728&0.818&\textbf{0.842}&0.752&0.922&1.123&0.368&0.407&0.463\\
TimeMixer++ & 0.268 &0.430&0.501&\underline{0.617}&\textbf{0.791}&1.318&0.335&\textbf{0.366}&0.589&\textbf{0.214}&0.342&0.456\\
\midrule
Timemachine & \textbf{0.249} &\underline{0.345}&\underline{0.479}&0.713&0.924&1.501&\textbf{0.328}&0.412&0.483&0.261&0.395&0.463\\
\midrule
CSDI        & 1.005 &1.063&1.077&0.682&0.836&1.138&0.787&1.015&1.104&0.306&0.525&0.586\\
D3VAE       & 0.669 &0.727&1.071&0.680&0.825&1.177&0.543&0.620&0.674&0.329&0.380&0.547\\
\midrule
FPT         & 0.416 &0.517&0.597&0.621&1.117&1.433&0.412&0.448&0.451&0.252&0.384&0.450\\
\midrule
TimeLLM     & 0.426 &0.524&0.528&0.712&0.864&1.187&0.362&0.389&0.440&0.242&0.344&\underline{0.422}\\
MM-TSF      & 0.711 &0.798&0.825&\textbf{0.608}&0.892&1.239&0.736&0.805&0.921&0.406&0.575&0.787\\
\midrule
GLAFF       & 0.518 &0.549&0.591&1.080&1.178&1.367&0.344&0.579&0.573&0.223&0.313&0.443\\
TimeLinear  & 0.459 &0.530&0.657&0.811&1.008&1.283&0.403&0.440&0.527&0.321&0.381&0.466\\
\midrule
\midrule
MCD-TSF     & \underline{0.251} &\textbf{0.323}&\textbf{0.392}&0.763&1.015&1.134&\underline{0.333}&\underline{0.380}&\underline{0.409}&\underline{0.220}&\underline{0.292}&\textbf{0.367}\\
\bottomrule[2pt]
\end{tabular}%
}
  \end{subtable}
  \vspace{0.1em}
  \begin{subtable}[t]{\textwidth}
    \centering
    \caption{MAE Results 2}
    \label{tab:comp_baselines_all_horizons_mse}
\resizebox{\columnwidth}{!}{
\begin{tabular}{c|ccc|ccc|ccc|ccc}
\toprule[2pt]
            & \multicolumn{3}{c|}{Environment} & \multicolumn{3}{c|}{Health} & \multicolumn{3}{c|}{Social Good} & \multicolumn{3}{c}{Traffic} \\
\midrule
            & 48  & 96  & 192 & 12  & 24  & 48  & 6   & 12  & 18  & 6   & 12  & 18  \\
\midrule
Reformer    & 0.435&0.449&0.461&0.853&0.862&0.866&0.760&0.879&0.906&0.439&0.494&0.566\\
Autoformer  & 0.415&0.443&0.456&0.859&0.860&0.888&0.586&0.753&0.836&0.255&0.271&0.285\\
FEDformer   & 0.392&0.419&0.425&0.874&0.948&1.007&0.797&0.953&1.028&0.283&0.296&0.298\\
PatchTST    & \underline{0.362}&\textbf{0.368}&\textbf{0.371}&0.786&0.868&0.937&0.445&0.570&0.714&0.201&0.235&0.253\\
HCAN        & 0.387&0.428&0.445&0.783&0.910&0.913&0.459&0.602&0.701&0.241&0.333&0.631\\
\midrule
FiLM        & \textbf{0.352}&0.377&0.436&1.013&1.198&1.219&0.931&1.016&1.193&0.263&0.272&0.341\\
DLinear     & 0.380&0.379&0.387&0.804&0.881&0.950&0.769&0.851&0.924&0.303&0.435&0.465\\
TimeMixer++ & 0.441&0.445&0.458&0.771&0.900&1.127&\underline{0.430}&0.564&0.773&\underline{0.149}&0.170&\underline{0.188}\\
\midrule
Timemachine & 0.370&0.384&0.388&0.777&0.846&1.012&0.494&0.570&0.755&0.161&\underline{0.172}&\underline{0.188}\\
\midrule
CSDI        & 0.378&0.389&0.406&0.735&0.890&0.948&0.487&0.594&\textbf{0.648}&0.232&0.241&0.243\\
D3VAE       & 0.472&0.558&0.577&0.868&0.946&\underline{0.867}&0.550&0.658&\underline{0.689}&0.355&0.374&0.378\\
\midrule
FPT         & 0.397&0.383&0.437&0.797&0.866&0.868&0.450&\underline{0.561}&0.749&0.310&0.318&0.324\\
\midrule
TimeLLM     & 0.445&\underline{0.370}&\underline{0.380}&\underline{0.704}&\underline{0.787}&0.964&0.554&0.628&0.712&0.224&0.227&0.234\\
MM-TSF      & 0.450&0.463&0.488&0.706&\textbf{0.767}&\textbf{0.809}&0.484&0.620&0.692&0.382&0.409&0.417\\
\midrule
GLAFF       & 0.378&0.380&0.438&0.750&0.875&0.918&\textbf{0.429}&0.586&0.714&0.198&0.203&0.217\\
TimeLinear  & 0.377&0.388&0.397&0.799&0.847&0.906&0.777&0.880&0.994&0.275&0.307&0.318\\
\midrule
\midrule
MCD-TSF     & 0.369&0.384&0.385&\textbf{0.651}&0.871&0.912&0.440&\textbf{0.531}&0.736&\textbf{0.138}&\textbf{0.164}&\textbf{0.174}\\
\bottomrule[2pt]
\end{tabular}%
}
  \end{subtable}
\end{table}

\subsection{Effect of the Timestamp Weight}
\label{app:timestamp-weight}

\begin{figure}[t]
    \centering
    \includegraphics[width=\linewidth]{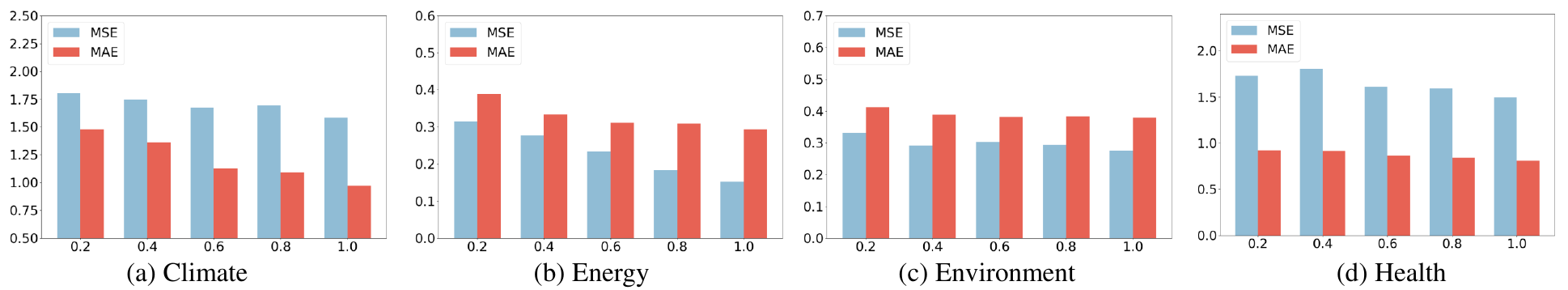}
    \caption{{Results of our approach on different domains when configured with various values of the timestamp weights (i.e., the $\lambda=0.2, 0.4, 0.6, 0.8, 1.0$).}}
    \label{fig:timestamp-weight-2}
\end{figure}

{
Figure \ref{fig:timestamp-weight-2} presents the results of our approach on \textit{climate}, \textit{energy}, \textit{environment}, and \textit{health} domains when configured with various values of the timestamp weights, where a similar trend to the one in Section \ref{sec:timestamp-effect} is observed.
}

\subsection{Effect of the Unconditional Training Probability}
\label{app:Unconditional-Training-Probability}

\begin{figure}[t]
    \centering
    \includegraphics[width=\linewidth]{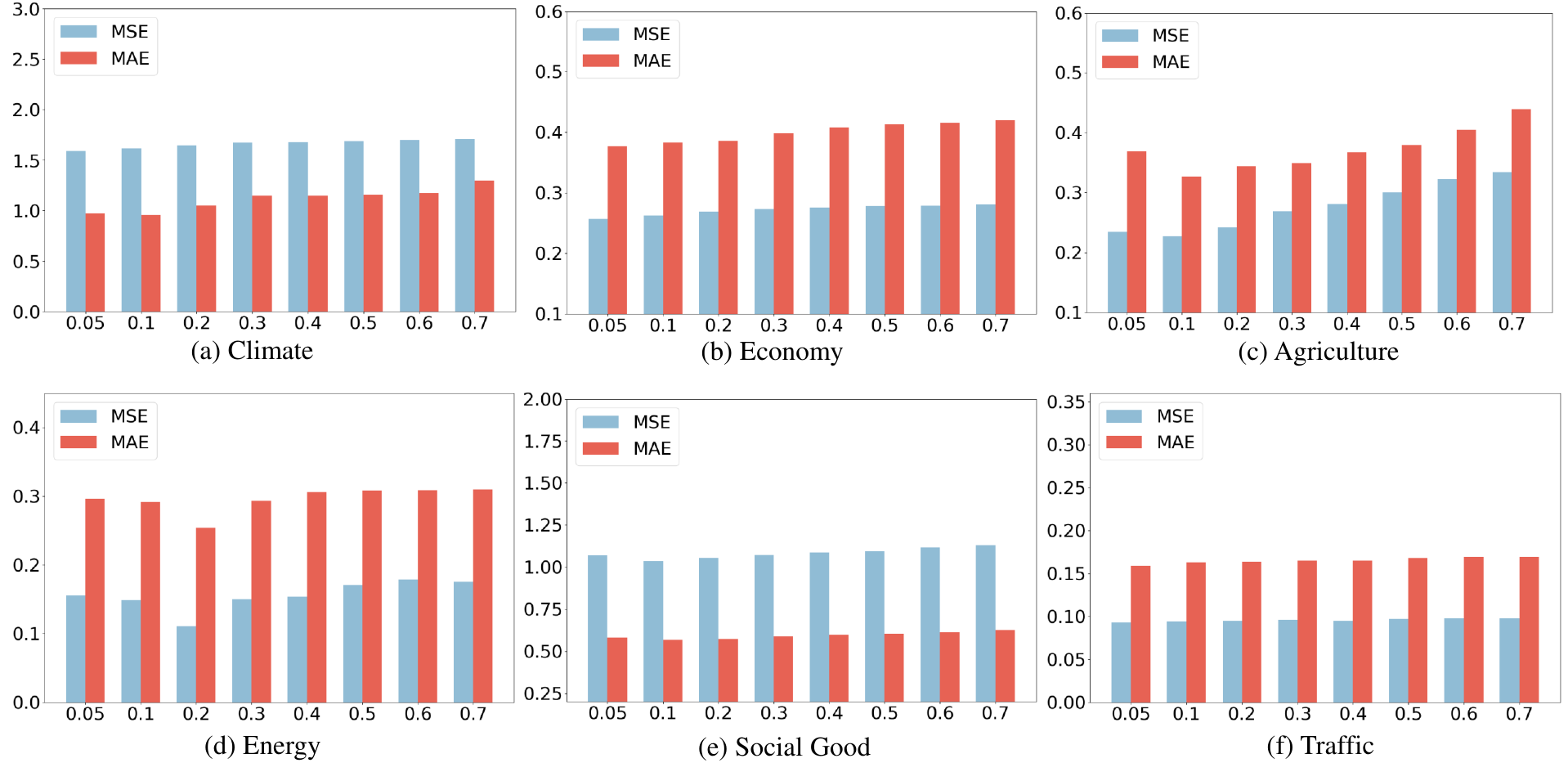}
    \caption{{The model performance with changes in the unconditional training probability.}}
    \label{fig:uncond_others}
\end{figure}

{
Figure \ref{fig:uncond_others} presents the results of our model with different values of the unconditional training probability \( p_{\text{uncond}} \) in various domains, namely, \textit{agriculture}, \textit{climate}, \textit{economy}, \textit{energy}, \textit{social good}, and \textit{traffic}.
The trend is similar to the observations in Section \ref{sec:text-effect}.
}

\end{document}